\documentclass{article}

\PassOptionsToPackage{square,numbers}{natbib}
\usepackage[final]{neurips_2021}

\usepackage[utf8]{inputenc} 
\usepackage[T1]{fontenc}    
\usepackage{hyperref}       
\usepackage{url}            
\usepackage{booktabs}       
\usepackage{amsfonts}       
\usepackage{nicefrac}       
\usepackage{microtype}      
\usepackage{xcolor}         
\usepackage{amsmath}
\usepackage{amssymb}
\usepackage{graphicx}
\usepackage{caption}
\usepackage{subcaption}
\usepackage{multirow}
\usepackage{xcolor}
\usepackage{amsthm}
\usepackage{comment}

\newtheorem{theorem}{Theorem}

\title{Contrastively Disentangled \\ Sequential Variational Autoencoder}

\author{%
  Junwen Bai \\
  Cornell University\\
  \texttt{jb2467@cornell.edu} \\
  \And
  Weiran Wang \\
  Google \\
  \texttt{weiranwang@google.com} \\
  \And
  Carla Gomes \\
  Cornell University \\
  \texttt{gomes@cs.cornell.edu} \\
}

\begin{document}

\maketitle

\begin{abstract}
Self-supervised disentangled representation learning is a critical task in sequence modeling. The learnt representations contribute to better model interpretability as well as the data generation, and improve the sample efficiency for downstream tasks. We propose a novel sequence representation learning method, named Contrastively Disentangled Sequential Variational Autoencoder (C-DSVAE), to extract and separate the static (time-invariant) and dynamic (time-variant) factors in the latent space. Different from previous sequential variational autoencoder methods, we use a novel evidence lower bound which maximizes the mutual information between the input and the latent factors, while penalizes the mutual information between the static and dynamic factors. We leverage contrastive estimations of the mutual information terms in training, together with simple yet effective augmentation techniques, to introduce additional inductive biases. Our experiments show that C-DSVAE significantly outperforms the previous state-of-the-art methods on multiple metrics. 
\end{abstract}

\section{Introduction}
\label{s:intro}

The goal of self-supervised learning methods is to extract useful and general representations without any supervision, and to further facilitate downstream tasks such as generation and prediction~\citep{bengio2013representation}. Despite the difficulty of this task, many existing works have shed light on this field across different domains such as computer vision~\citep{oord2018representation, hjelm2018learning,chen2020simple,golinski2020feedback},
natural language processing~\citep{peters2018deep,devlin2018bert,brown2020language} and speech processing~\citep{baevski2019effectiveness,chung2020generative,wang2020unsupervised, baevski2020wav2vec, bai2021representation} (also see a huge number of references in these papers). While the quality of the learnt representations improves gradually, recent research starts to put more emphasis on learning disentangled representations. This is because disentangled latent variables may capture separate variations of the data generation process, which could contain semantic meanings, provide the opportunity to remove unwanted variations for a lower sample complexity of downstream learning \citep{villegas2017decomposing, denton2017unsupervised}, and allow more controllable generations \citep{zhu2018generative,tian2021good,han2021disentangled}. These advantages lead to a rapidly growing research area, studying various principles and algorithmic techniques for disentangled representation learning~\citep{chen2016infogan, higgins2016beta,fraccaro2017disentangled,kim2018disentangling,chen2018isolating,locatello2019challenging,khemakhem2020variational,locatello20a}. 
One concern raised in~\citep{locatello2019challenging} is that without any inductive bias, it would be extremely hard to learn meaningful disentangled representations. On the other hand, this concern could be much alleviated in the scenarios where the known structure of the data can be exploited.

In this work, we are concerned with the representation learning for sequence data, which has a unique structure to utilize for disentanglement learning. 
More specifically, for many sequence data, the variations can be explained by a dichotomy of a static (time-invariant) factor and dynamic (time-variant) factors, 
each varies independently from the other.
For example, representations of a video recording the movements of a cartoon character 
could be disentangled into the character identity (static) and the actions (dynamic). For audio data, the representations shall be able to separate the speaker information (static) from the linguistic information (dynamic).

We propose Contrastively Disentangled Sequential Variational Autoencoder (C-DSVAE), a method seeking for a clean separation of the static and dynamic factors for the sequence data. Our method extends the previously proposed sequential variational autoencoder (VAE) framework, and performs learning with a different evidence lower bound (ELBO) which naturally contains mutual information (MI) terms to encourage disentanglement. Due to the difficulty in estimating high dimensional complex distributions (e.g., for the dynamic factors), we further incorporate the contrastive estimation for the MI terms with systematic data augmentation techniques which modify either the static or dynamic factors of the input sequence. The new estimation method turns out to be more effective than the minibatch sampling based estimate, and introduces additional inductive biases towards the invariance. To our knowledge, we are the first to synergistically combine the contrastive estimation and sequential generative models in a principled manner for learning disentangled representations.
We validate C-DSVAE on four datasets from the video and audio domains. The experimental results show that our method consistently outperforms the previous state-of-the-art (SOTA) methods, both quantitatively and qualitatively.

\section{Method}
\label{s:method}

We denote the observed input sequence as $x_{1:T}=\{x_1, x_2, ..., x_T\}$ where $x_i$ represents the input feature at time step $i$ (e.g., $x_i$ could be a video frame or the spectrogram feature of a short audio segment), and $T$ is the sequence length. The latent representations are divided into the static factor $s$, and the dynamic factors $z_{1:T}$ where $z_i$ is the learnt dynamic representation at time step $i$. 

\subsection{Probabilistic Model}
\label{ss:probabilistic}

\begin{figure}[t]
    \includegraphics[width=0.97\textwidth]{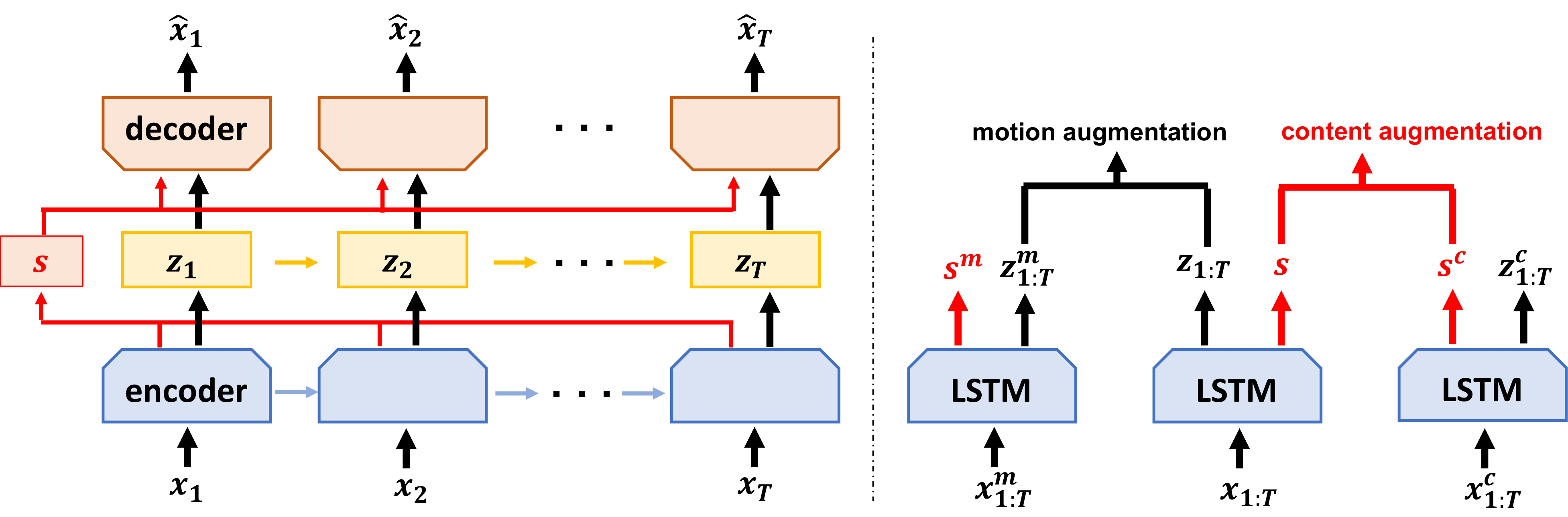}
    \caption{The illustration of our C-DSVAE model. \textbf{Left panel} is the general structure of the sequence-to-sequence auto-encoding process: each frame is passed to the LSTM cell; dynamic factors $z_{1:T}$ are extracted for each time step; the static factor $s$ is extracted by summarizing the full sequence; the generation/reconstruction of frame $i$ depends on $s$ and $z_i$. \textbf{Right panel} depicts the contrastive learning module of C-DSVAE: $x_{1:T}^m$ is the motion augmentation of $x_{1:T}$ and $x_{1:T}^c$ is the content augmentation of $x_{1:T}$;  dynamic factors $z_{1:T}^m$ of $x_{1:T}^m$ can be seen as the positive sample for the anchor $z_{1:T}$ in the contrastive estimation w.r.t. the motion, and similarly the static factor $s^c$ of $x_{1:T}^c$ can be viewed as the positive sample for $s$ w.r.t. the content. 
    }
    \label{fig:cdsvae}
\end{figure}

We assume that in the ground-truth generation process, $z_i$ depends on $z_{<i}=\{z_0, z_1, ..., z_{i-1}\}$ where $z_0=\mathbf{0}$, and the observation $x_i$ is independent of other frames conditioned on $z_i$ and $s$. Furthermore, we assume the static variable $s$ and the dynamic variables $z_{1:T}$ are independent from each other, i.e., $p(s, z_{1:T})=p(s)p(z_{1:T})$.  Formally, let $z=(s, z_{1:T})$ and we have the following complete likelihood
\begin{equation}
    p(x_{1:T}, z)=p(z)p(x_{1:T}|z)=\bigg[p(s)\prod_{i=1}^T p(z_i|z_{<i})\bigg] \cdot \prod_{i=1}^T p(x_i|z_i, s)
\label{eq:generation_process}
\end{equation}
where $p(z)$ is the prior for sampling $z$. 
Our formulation captures the general intuition that we can separate the variations of the sequence into the time-dependent dynamic component (described by $z_i$'s), and a static component (described by $s$) which remains the same for different time steps in the same sequence but differs between sequences. For example, in the cartoon character video (see Figure~\ref{fig:sprites}), the hair, shirt and pants should be kept the same when the character walks around, but different videos can have different characters. Similarly, for a speech utterance, the phonetic transcription controls the vocal tract motion and the sound being produced over time, but the speaker identity remains the same for the whole utterance. 
In this work, we refer to the dynamic component as "motion" and the static component as "content".

For the prior distributions, we choose $p(s)$ to be the standard Gaussian $\mathcal N(0, I)$, and $p(z_i|z_{<i})$ to be $\mathcal N(\mu(z_{<i}), \sigma^2(z_{<i}))$ where $\mu(\cdot)$ and $\sigma(\cdot)$ are modeled by LSTMs \citep{hochreiter1997long}. In operations, the drawn sample $s$ is shared throughout the sequence. To draw a sample of $z_t$, we first take a sample of $z_{t-1}$ as the input of the current LSTM cell. Then forwarding the LSTM for one step gives us the distribution of $z_t$, from which we draw a sample with the reparameterization trick \citep{kingma2014auto}.

To extract the latent representations given only the observed data $x_{1:T}$ where the motion and content are mixed together, we hope to learn a posterior distribution $q(z|x_{1:T})$ 
where the two components are disentangled. That is, similar to the prior, our posterior should have a factorized form:
\begin{equation}
\begin{aligned}
    q(z|x_{1:T})&=q(z_{1:T}, s|x_{1:T})=q(z_{1:T}|x_{1:T})q(s|x_{1:T})=q(s|x_{1:T})\prod_{i=1}^T q(z_i|z_{<i}, x_{\le i}).
\end{aligned}
\end{equation}
The posterior distributions are also modeled by LSTMs.
A nested sampling procedure, resembling that of the prior, is applied to the posterior of dynamic variables. Similar parameterizations of dynamic variables by recurrent networks have been proposed in prior works \citep{chung2015recurrent, goyal2017z, krishnan2015deep}.
The standard loss function for learning the latent representations is an ELBO \citep{li2018disentangled,zhu2020s3vae} (also see a derivation in Appendix~\ref{ss:derivation_of_per_seq_elbo}):
\begin{equation}
\label{eq:dsvae}
\begin{aligned}
\max_{p,q} \; \mathbb E_{x_{1:T}\sim p_D} \mathbb E_{q(z|x_{1:T})} \left[\log p(x_{1:T}|z)- KL[q(z|x_{1:T})||p(z)] \right]
\end{aligned}
\end{equation}
where $p_D$ is the empirical data distribution. 
Under the parameterization of the posterior where $s$ and $z_{1:T}$ are mutually independent, the KL-divergence term reduces to
\begin{equation}
\label{eq:kl}
\begin{aligned}
KL[q(z|x_{1:T})||p(z)]=KL[q(s|x_{1:T})||p(s)] + KL[q(z_{1:T}|x_{1:T})||p(z_{1:T})]
\end{aligned}
\end{equation}
where the second term is approximated with the sampled trajectories of the dynamic variables $z_{1:T}$.
 

\subsection{Our approach: Mutual information-based disentanglement}
\label{ss:cdsvae}
Several existing sequence representation learning methods \citep{li2018disentangled,zhu2020s3vae,han2021disentangled} are built on top of the loss function \eqref{eq:dsvae}.
However, this formulation also brings several issues. The KL-divergence regularizes the posterior of the static or dynamic factors to be close to the corresponding priors. When modeling them with powerful neural architectures like LSTMs, it is possible for the KL-divergence to be close to zero, yet at the same time, the posteriors become non-informative of the inputs; this is a common issue for deep generative models like VAE~\citep{bowman2015generating}. 
Techniques have been proposed to alleviate this issue, including adjusting the relative weights of the loss terms to regularize the capacity of posteriors \citep{higgins2016beta}, replacing the individual posteriors in the KL terms with aggregated posteriors \citep{tomczak2018vae,alemi2018fixing,takahashi2019variational}
and enforcing latent structures (such as disentangled representations) in the posteriors~\citep{chen2018isolating,kim2018disentangling,tschannen2018recent}.

Our principled approach for learning useful representations from sequences is inspired by these prior works, and at the same time incorporates the unique sequential structure of the data.
Without inductive biases, the goal of disentanglement can hardly be achieved since it is possible to find entangled $s$ and $z_{1:T}$ that explain the data equally well (in fact, by Theorem 1 of~\citep{locatello2019challenging}, there could exist infinitely many such entangled factors). The problem may seem less severe in our setup, since $s$ is shared across time and it is hard for such $s$ to capture all dynamics.
Nevertheless, since both $s$ and $z_i$ are used in generating $x_i$, it is still possible for $z_i$ to carry some static information. In the extreme case where each $z_i$ encompasses $s$, $s$ would no longer be indispensable for the generation. 
This issue motivated prior works to optimize the (estimate of) mutual information among latent variables and inputs~\citep{zhu2020s3vae,han2021disentangled,akuzawa2021information}.




Our method seeks to achieve clean disentanglement of $s$ and $z_{1:T}$, by optimizing the following objective function, which introduces additional MI terms to the vanilla ELBO in~\eqref{eq:dsvae}:
\begin{align}
\max_{p,q}\; &\mathbb E_{x_{1:T}\sim p_D}\mathbb E_{q(z|x_{1:T})}[\log p(x_{1:T}|z)]-KL[q(z)||p(z)] \nonumber \\
=\; & \mathbb E_{x_{1:T}\sim p_D}\mathbb E_{q(z|x_{1:T})}[\log p(x_{1:T}|z)] - (KL[q(s|x_{1:T})||p(s)]+KL[q(z_{1:T}|x_{1:T})||p(z_{1:T})]) \nonumber \\ \label{eq:dsvae_mi}
&\hspace{4em} + I_q (s;x_{1:T}) + I_q (z_{1:T};x_{1:T}) - I_q (z_{1:T};s)
\end{align}
where the aggregated posteriors are defined as
$q(z)=q(s,z_{1:T})=\mathbb E_{p_D} [q(s|x_{1:T}) q(z_{1:T}|x_{1:T})]$, $q(s)=\mathbb E_{p_D}[q(s|x_{1:T})]$, $q(z_{1:T})=\mathbb E_{p_D}[q(z_{1:T}|x_{1:T})]$, and the MI terms are defined as $I_q (s;x_{1:T})=\mathbb E_{q(s, x_{1:T})} \left[\log \frac{q(s|x_{1:T})}{q(s)}\right]$, $I_q (z_{1:T};x_{1:T})=\mathbb E_{q(z_{1:T}, x_{1:T})} \left[\log \frac{q(z_{1:T}|x_{1:T})}{q(z_{1:T})}\right]$ (and $I_q (z_{1:T};s)$ is defined similarly).
The intuition behind~\eqref{eq:dsvae_mi} is simple: besides explaining the data and matching the posteriors with priors, $z_{1:T}$ and $s$ shall contain useful information from $x_{1:T}$ while excluding the redundant information from each other.
We further justify~\eqref{eq:dsvae_mi} by showing that it still forms a valid ELBO.
\begin{theorem}
\label{thm1}
With our parameterization of $(s,z_{1:T})$, \eqref{eq:dsvae_mi} is a valid lower bound of the data log-likelihood $\mathbb E_{x_{1:T}\sim p_D} \log (x_{1:T})$.
\end{theorem}
The full proof can be found in Appendix~\ref{ss:thm1_proof}. With this guarantee, we can follow the spirit of~\citep{higgins2016beta, chen2018isolating} to add and adjust the additional weight coefficients $\alpha$ to the KL terms, $\beta$ to the MI terms $I_q(s;x_{1:T})$, $I_q(z_{1:T};x_{1:T})$, and $\gamma$ to $I_q (z_{1:T};s)$,
\begin{equation}
\label{eq:dsvae_mi_final}
\begin{aligned}
\mathbb E_{x_{1:T}\sim p_D}&\mathbb E_{q(z|x_{1:T})}[\log p(x_{1:T}|z)]-\alpha(KL[q(s|x_{1:T})||p(s)]+KL[q(z_{1:T}|x_{1:T})||p(z_{1:T})])\\
&+\beta(I_q(s;x_{1:T})+I_q(z_{1:T};x_{1:T}))-\gamma I_q(z_{1:T};s).
\end{aligned}
\end{equation}
It remains to estimate the objective~\eqref{eq:dsvae_mi_final} for optimization.
The KL term for $z_{1:T}$ is estimated with the standard Monte-Carlo sampling, using trajectories of $z_{1:T}$ \citep{li2018disentangled}. The KL term for $s$ can be estimated analytically.
For the MI terms, we attempt two estimations. The first estimation uses a standard mini-batch weighted sampling (MWS) following~\citep{chen2018isolating,zhu2020s3vae}. Due to the high dimensionality of $z_{1:T}$ and the complex dependency among time steps, it may be hard for MWS to estimate the distributions accurately, so we also explore non-parametric contrastive estimations for $I_q(s;x_{1:T})$ and $I_q(z_{1:T};x_{1:T})$ with additional data augmentations, which we will detail below.

\subsection{C-DSVAE: Contrastive estimation with augmentation} 
\label{ss:augmentation}

A contrastive estimation of $I(z_{1:T};x_{1:T})$ can be defined as follows
\begin{equation}
\begin{aligned}
\mathcal C(z_{1:T}) = \mathbb E_{p_D} \log \frac{\phi(z_{1:T}, x_{1:T}^+)}{\phi(z_{1:T}, x_{1:T}^+) + \sum_{j=1}^n \phi(z_{1:T}, x_{1:T}^j)} + \log (n+1)
\end{aligned}
\label{eq:contrast}
\end{equation}
where $x^+$ is a "positive" sequence, while $x^j$, $j=1,\dots,n$ is a collection of $n$ "negative" sequences. When $\phi(z_{1:T}, x_{1:T})=\frac{q(x_{1:T}|z_{1:T})}{q(x_{1:T})}$, one can show that~\eqref{eq:contrast} approximates  
$I_q(z_{1:T},x_{1:T})$; see Appendix~\ref{ss:contrast_est_mi} for a proof.
To implement~\eqref{eq:contrast}$, z_{1:T}$ is the trajectory obtained by using the mean at each time step from $q(z_{1:T}|x_{1:T})$ for a input sequence $x_{1:T}$, while $x_{1:T}^j$ is a randomly sampled sequence from the minibatch. A similar contrastive estimation is defined for $s$ as well.
To provide meaningful positive sequences that encourage the invariance of the learnt representations, we obtain $x_{1:T}^+$ by systematically perturbing $x_{1:T}$. 
In Sec~\ref{ss:contrast_est_mi}, we discuss how the augmented data give good estimate of~\eqref{eq:contrast} under additional assumptions.

\paragraph{Content augmentation} 
The static factor (e.g., the character identity in videos or the speaker in audios) is shared across all the time steps, and should not be affected by the exact order of the frames. We therefore randomly shuffle or simply reverse the order of time steps to generate the content augmentation of $x_{1:T}$ and denote it by $x_{1:T}^c$. The static and dynamic latent factors of $x_{1:T}^c$, modeled by $q$, are denoted by $s^c$ and $z_{1:T}^c$. 
This is an inexpensive yet useful strategy, applicable to both audios and videos. Similar ideas were introduced in \citep{zhu2020s3vae} albeit not used for contrastive estimations. 

\begin{figure}[t]
  \begin{subfigure}{0.495\textwidth}
    \includegraphics[width=\textwidth]{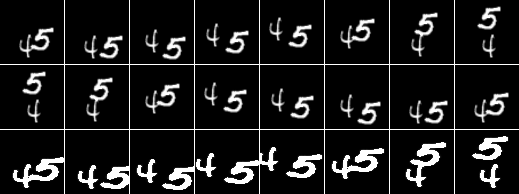}
    \label{fig:smmnist_aug}
  \end{subfigure}
  \hfill
  \begin{subfigure}{0.495\textwidth}
    \includegraphics[width=\textwidth]{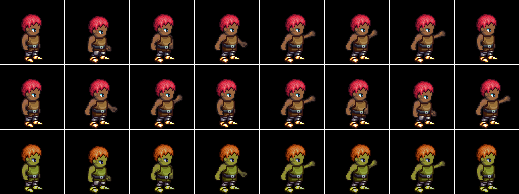}
    \label{fig:sprite_aug}
  \end{subfigure}
  \caption{Data augmentations on SM-MNIST and Sprites. \textbf{Left panel}: SM-MNIST. The first row is the raw input sequence of moving digits. The second row reverses the sequence order of the raw sequence, serving as the content augmentation. The third row stretches the frames and enhances the color, which changes the digit styles but does not change the digit movements and thus forms a motion augmentation. \textbf{Right panel}: Sprites. The first row is the raw input character video. The second row is the content augmentation with a random order. The third row is a motion augmentation produced by distorting colors and adding Gaussian noise.}
  \label{fig:aug}
\end{figure}

\paragraph{Motion augmentation}
In motion augmentation, we would like to maintain the dynamic factors (e.g., actions or movements) while replacing the content with a meaningful alternative. Thanks to the recent efforts in contrastive learning, multiple effective strategies have been proposed. 
For video datasets, we adopt the combination of cropping, color distortion, Gaussian blur and reshaping~\citep{chen2020simple,bachman2019learning}. For audio datasets, we use classical unsupervised voice conversion algorithms
~\citep{verhelst1993overlap,driedger2016review}. The motion augmentation of $x_{1:T}$ is denoted by $x_{1:T}^m$, with the static factor $s^m$ and dynamic factors $z_{1:T}^m$ estimated by $q$.

Note that $x_{1:T}^m$ is the motion augmentation of $x_{1:T}$, and $x_{1:T}$ in turn is also the motion augmentation of $x_{1:T}^m$. Similarly, $s$ and $s^c$ are mutually the "positive" sample to each other w.r.t. the static factor. 
During the training, it is efficient to generate the augmentations for each sequence in the minibatch, and apply contrastive estimation on both the original data and the augmented data. This leads to the following final estimates:
\begin{equation}
\label{eq:mi_est}
\begin{aligned}
I_q(z_{1:T};x_{1:T}) \approx \frac{1}{2}(\mathcal C(z_{1:T}) + \mathcal C(z_{1:T}^m)), \qquad 
I_q(s;x_{1:T}) \approx \frac{1}{2}(\mathcal C(s)+\mathcal C(s^c))
\end{aligned}
\end{equation}
where we use $\phi(z_{1:T}(x_{1:T}), x_{1:T}^*) =  \exp ( \text{sim}(z_{1:T}, z_{1:T}^*) /\tau )$ with $^*$ indicating the latent variables for the positive/augmented or negative sequences (extracted from $q$). $\text{sim}(\cdot,\cdot)$ is the cosine similarity function and $\tau=0.5$ is a temperature parameter.
This form of $\phi$ using cosine similarity is widely adopted in contrastive learning \citep{chen2020simple,khosla2020supervised} as it removes one degree of freedom (length) for high dimensional feature space.
Plugging~\eqref{eq:mi_est} into \eqref{eq:dsvae_mi_final} gives our final learning objective with contrastive estimation.
We name our method Contrastively Disentangled Sequential Variational Encoder (C-DSVAE), and a full model illustration is shown in Figure~\ref{fig:cdsvae}. 

In practice, we find the contrastive estimation is more effective than the MWS based on Gaussian probabilities (see an ablation study in Appendix~\ref{s:ablation}). Interestingly, we observe that with only the contrastive estimation and augmentations, $I_q(z_{1:T}; s)$ (last term in~\eqref{eq:dsvae_mi}) decreases even if we do not include its estimation in our optimization process (see Appendix~\ref{ss:mi_observations}), indicating that a good disentanglement between $z_{1:T}$ and $s$ could be attained through contrastive estimations solely.

As it will become evident in the experiments, our method achieves a cleaner separation of the dynamic and static factors than previous methods. We believe this is partly due to the inductive bias introduced by our approach: in order to consistently map two sequences with the same motion but different contents (which vary independently from the motion according to the generation process~\eqref{eq:generation_process}) to the same latent representation $z_{1:T}$, we must discard the information of the inputs regarding $s$; similar arguments hold for the static variables. At the same time, we enforce that the extracted $z_{1:T}$ and $s$ together should reconstruct $x_{1:T}$ so that no information is lost in the auto-encoding process.
To our knowledge, we are the first to use the contrastive estimation with augmentations in the sequential VAE scenario, for pushing the disentanglement of variations.

\section{Related Work}
\label{s:related}

In terms of the general frameworks, disentangled representation learning methods can be categorized into GAN-based~\citep{NIPS2014_Ian} and VAE-based~\citep{kingma2014auto} approaches. 
GAN-based methods like MoCoGAN~\citep{tulyakov2018mocogan} aim at generating videos from content noise and motion noise. These models typically do not explicitly learn the encoder from inputs to the latent variables, rendering them less applicable for representation learning.

The VAE framework parameterizes both the encoder (variational posterior) and the decoder (reconstruction), and optimize them jointly with a well-defined ELBO objective. The encoder allows us to directly control and reason about the properties of the extracted features, which can be straightforwardly used for downstream tasks. 
Many variants of VAE were proposed to encourage the disentanglement/interpretability of \emph{individual latent dimensions}. $\beta$-VAE \citep{higgins2016beta} adds a coefficient to the per-sample KL-divergence term to better constrain the information bottleneck. FactorVAE \citep{kim2018disentangling} and $\beta$-TCVAE \citep{chen2018isolating} explicitly separate out the total correlation term, and adjust its weight coefficient in the objective to encourage the per-dimensional disentanglement of the \emph{aggregated} posterior.
From a theoretical standpoint, \citep{locatello2019challenging} points out that without further inductive biases, it is impossible to learn disentangled representations with the standard VAE. 

To handle the sequence data, vanilla VAE is extended to the recurrent version, and prior works have explored different approaches for separating the content and the motion.
FHVAE~\citep{hsu2017unsupervised} designs a hierarchical VAE model which uses "sequence-dependent variables" (corresponding to the speaker factor) and "sequence-independent variables" (corresponding to the linguistic factors) for modeling speech sequences, and performs learning with the per-sequence ELBO while respecting the hierarchy in designing the posterior. FHVAE does not use additional loss terms for encouraging the disentanglement.
Different from FHVAE, DSVAE~\citep{li2018disentangled} explicitly models the static and dynamic factors in its graphical model and its posterior has a factorized form.
A few recent works (including ours) inherit the clean formulation of DSVAE in terms of the graphical model and prior/posterior parameterization. S3VAE~\citep{zhu2020s3vae} introduces additional loss terms to the objective of DSVAE in an ad-hoc fashion, such as a triplet loss that encourages the invariance of the learnt static factors by permuting the frame order (spiritually similar to our contrastive estimation for $I(s,x_{1:T})$), an additional prediction loss on $z_{1:T}$ leveraging external supervision for the motion labels, and an MI term of $I(s,z_{1:T})$ to be minimized. Our C-DSVAE differs from S3VAE in that we naturally introduce the MI terms from the fundamental principle of VAE, and use augmentations instead of external supervision for learning the static and dynamic factors. 
Also based on the DSVAE formulation, R-WAE~\citep{han2021disentangled} proposes to replace the distance measure between the aggregated posterior and the prior with the Wasserstein distance, in the belief that the KL divergence is too restrictive. 
In R-WAE, Maximum Mean Discrepancy (MMD) based estimation or GAN-based estimation of Wasserstein distance is used, either of which requires heavy tuning of hyperparameters. Another similar method, IDEL \citep{cheng2020improving}, optimizes the additional set of MI terms similar to ours. However, in our work, we show that these MI terms can be naturally derived from the fundamental principle of VAE, and we additionally propose the contrastive estimation and data augmentations to strengthen them, which was not done before.


Another relevant research area is nonlinear ICA, which tries to understand the conditions under which the disentanglement can be achieved, possibly with contrastive learning or VAE ~\cite{NIPS2016_d305281f,hyvarinen17a,gresele2019incomplete,khemakhem2020variational,locatello20a}. Our general approach is well-aligned with the current understanding in this direction. That is, while general disentanglement (per-dimensional disentanglement) is hard to achieve without stringent assumptions on the data generation process, we can seek some certain level of disentanglement (in our case, the group-wise disentanglement between the static and dynamic factors) given auxiliary information (which we provide through augmentations). Our intuitive arguments for the reason of separation (see the last paragraph of Sec~\ref{ss:augmentation}) show that VAE and contrastive learning may be complementary to each other, and combining them could potentially gain better disentanglement.

\section{Experiments}
\label{s:exp}

We compare C-DSVAE with the state-of-the-art sequence disentanglement learning methods: FHVAE~\citep{hsu2017unsupervised}, MoCoGAN~\citep{tulyakov2018mocogan}, DSVAE~\citep{li2018disentangled}, S3VAE~\citep{zhu2020s3vae} and R-WAE~\citep{han2021disentangled}, with the same experimental setups used by them.


\subsection{Datasets}

\textbf{Sprites}~\citep{reed2015deep} is a cartoon character video dataset. Each character's (dynamic) motion can be categorized into three actions (walking, spellcasting, slashing) and three directions (left, front, right). The (static) content comprises of each character's skin color, tops color, pants color and hair color. Each color has six variants. Every sequence is composed of 8 frames of RGB images with size $64\times 64$.

\textbf{MUG}~\citep{aifanti2010mug} is a facial expression video dataset. The static factor corresponds to a single person's identity. Each individual performs six expressions (motion): anger, fear, disgust, happiness, sadness and surprise. Following \citep{tulyakov2018mocogan}, each sequence contains 15 frames of RGB images with size $64\times 64$ (after resizing). 

\textbf{SM-MNIST} (\citep{denton2018stochastic}, Stochastic Moving MNIST) is a dataset that records the random movements of two digits. Each sequence contains 15 gray scale images of size $64\times 64$. Individual digits are collected from MNIST. 


\textbf{TIMIT}~\citep{timit1992} is a corpus of read speech for acoustic-phonetic studies and speech recognition. The utterances are produced by American speakers of eight major dialects reading phonetically rich sentences.
Following \citep{hsu2017unsupervised,han2021disentangled}, we extract per-frame spectrogram features (with a shift size of 10ms) from audio, 
and segments of 200ms duration (20 frames) are chunked from the original utterances and then treated as independent sequences for learning.
All datasets are separated into training, validation and testing splits following~\citep{hsu2017unsupervised,zhu2020s3vae,han2021disentangled}.

\begin{table}[t]
\begin{minipage}{.5\linewidth}
\centering
\caption{Disentanglement metrics on Sprites.}
\label{tab:sprites}
\begin{tabular}{c@{\hspace{0.4em}}c@{\hspace{0.4em}}c@{\hspace{0.4em}}c@{\hspace{0.4em}}c@{\hspace{0.4em}}}
\toprule
Methods & Acc$\uparrow$ & IS$\uparrow$ & H(y|x)$\downarrow$ & H(y)$\uparrow$ \\
\midrule
MoCoGAN & 92.89\% & 8.461 & 0.090 & 2.192 \\
DSVAE & 90.73\% & 8.384 & 0.072 & 2.192 \\
R-WAE & 98.98\% & 8.516 & 0.055 & 2.197 \\
S3VAE & 99.49\% & 8.637 & 0.041 & 2.197 \\
\midrule
C-DSVAE & \textbf{99.99\%} & \textbf{8.871} & \textbf{0.014} & \textbf{2.197} \\
\bottomrule
\end{tabular}
\end{minipage}%
\begin{minipage}{.5\linewidth}
\centering
\caption{Disentanglement metrics on MUG.}
\label{tab:mug}
\begin{tabular}{c@{\hspace{0.4em}}c@{\hspace{0.4em}}c@{\hspace{0.4em}}c@{\hspace{0.4em}}c@{\hspace{0.4em}}}
\toprule
Methods & Acc$\uparrow$ & IS$\uparrow$ & H(y|x)$\downarrow$ & H(y)$\uparrow$ \\
\midrule
MoCoGAN & 63.12\% & 4.332 & 0.183 & 1.721 \\
DSVAE & 54.29\% & 3.608 & 0.374 & 1.657 \\
R-WAE & 71.25\% & 5.149 & 0.131 & 1.771 \\
S3VAE & 70.51\% & 5.136 & 0.135 & 1.760 \\
\midrule
C-DSVAE & \textbf{81.16\%} & \textbf{5.341} & \textbf{0.092} & \textbf{1.775} \\
\bottomrule
\end{tabular}
\end{minipage} 
\end{table}

\begin{table}[t]
\begin{minipage}{.5\linewidth}
\centering
\caption{Disentanglement metrics on SM-MNIST.}
\label{tab:smmnist}
\begin{tabular}{c@{\hspace{0.4em}}c@{\hspace{0.4em}}c@{\hspace{0.4em}}c@{\hspace{0.4em}}c@{\hspace{0.4em}}}
\toprule
Methods & Acc$\uparrow$ & IS$\uparrow$ & H(y|x)$\downarrow$ & H(y)$\uparrow$ \\
\midrule
MoCoGAN & 74.55\% & 4.078 & 0.194 & 0.191 \\
DSVAE & 88.19\% & 6.210 & 0.185 & 2.011 \\
R-WAE & 94.65\% & 6.940 & 0.163 & 2.147 \\
S3VAE & 95.09\% & 7.072 & 0.150 & 2.106 \\
\midrule
C-DSVAE & \textbf{97.84\%} & \textbf{7.163} & \textbf{0.145} & \textbf{2.176} \\
\bottomrule
\end{tabular}
\end{minipage}%
\begin{minipage}{.5\linewidth}
\centering
\caption{Disentanglement metrics on TIMIT.}
\label{tab:timit}
\begin{tabular}{c@{\hspace{0.4em}}c@{\hspace{0.5em}}c@{\hspace{0.5em}}}
\toprule
Methods &  content EER$\downarrow$ & motion EER$\uparrow$ \\ 
\midrule
FHVAE & 5.06\% & 22.77\% \\ 
DSVAE & 5.64\% & 19.20\% \\ 
R-WAE & 4.73\% & 23.41\% \\ 
S3VAE & 5.02\% & 25.51\% \\ 
\midrule
C-DSVAE & \textbf{4.03\%} & \textbf{31.81\%} \\ 
\bottomrule
\end{tabular}
\end{minipage} 
\end{table}

\subsection{Disentanglement Metrics}
\label{ss:metrics}

The quantitative performance measures all the models from two aspects: first, by fixing either the static or dynamic factors and randomly sample the other, how well the fixed factor can be recognized; and second, with the randomly sampled factor, how different the generated sequence is from the original one. To this end, we pretrain a separate classifier $C$ to identify the static or dynamic factors (if available). The classifier $C$ is carefully trained with the full supervision and thus qualifies as a judge. 

We use five metrics to evaluate the disentanglement performance: accuracy, Inception Score, inter-entropy, intra-entropy and equal error rate. 
Accuracy (Acc) measures how well the fixed factor can be identified by the classifier $C$.
Inception Score ($IS$) \citep{zhou2017inception} computes the KL-divergence between the conditional predicted label distribution $p(y|x_{1:T})$ and the marginal predicted label distribution $p(y)$ from $C$.
Inter-Entropy is similar to IS, but measures the diversity solely on the marginal predicted label distribution $p(y)$; a higher inter-entropy indicates that the generated sequences are more diverse.
Intra-Entropy measures the entropy over $p(y|x)$; a lower intra-entropy means more confident predictions. 
Equal error rate (EER) \citep{dehak2010front} is used in the TIMIT experiment to evaluate the speaker identification accuracy.



\subsection{Hyperparameters and Architectures}

As is commonly done in VAE learning~\citep{higgins2016beta},
in~\eqref{eq:dsvae_mi_final} we add coefficients $\alpha$ to the KL terms, $\beta$ to the contrastive terms, while the coefficient $\gamma$ of $I(s;z_{1:T})$ is fixed to be 1. In our experiments, $\alpha$ is tuned over $\{0.6, 0.9, 1.0, 2.0\}$ and $\beta$ is tuned over $\{0.1, 0.2, 0.5, 0.7, 1.0, 2.0, 5.0\}$. We use the Adam optimizer \citep{kingma2014adam} with the learning rate chosen from $\{0.0005, 0.001, 0.0015, 0.002\}$ through grid search. 
Models are trained until convergence and the one with the best validation accuracy would be chosen for testing.
Our network architecture follows~\citep{zhu2020s3vae}: motion priors and posteriors are both parameterized by uni-directional LSTMs, which output $\mu$ and $\sigma$ at each time step for the dynamic factors, while the content posterior produces $\mu$ and $\sigma$ for the static factor of the whole sequence with a bi-directional LSTM. See Appendix~\ref{ss:network_architecture} for further details.

\subsection{Quantitative Results}
\label{ref:quan}

Tables \ref{tab:sprites}, \ref{tab:mug}, \ref{tab:smmnist}, \ref{tab:timit} demonstrate the disentanglement evaluations on the 
datasets; $\uparrow$ means the higher the value the better the performance, and $\downarrow$ means the reverse. 
In Table~\ref{tab:sprites}, we fix 
$z_{1:T}$ and randomly sample $s$ from $p(s)$ for the evaluation. We observe that R-WAE, S3VAE, C-DSVAE can all deliver high accuracies in predicting a total of 9 classes (3 actions $\times$ 3 directions), but C-DSVAE outperforms all the others. 
On the other hand, when we fix $s$ and randomly sample $z_{1:T}$, all the methods achieve near-perfect accuracies for predicting character identities (and thus we do not show this result here). 
Similarly, in Table~\ref{tab:mug}, we evaluate the ability to maintain the facial expression by sampling $s$ and fixing $z_{1:T}$. C-DSVAE outperforms R-WAE and S3VAE by over 10\% relatively in the prediction accuracy, and its intra-entropy is lower by 30\% relatively. Table \ref{tab:smmnist} measures how well the static factor $s$, which corresponds to one of the total 100 digit combinations, can be recognized when the dynamic factors are randomly sampled (the dynamic factors are continuous and thus hard to be categorized). While R-WAE, S3VAE, and C-DSVAE can all generate high-quality moving digits (see Sec~\ref{ss:qual}), C-DSVAE non-trivially outperforms the others on all metrics. 

For TIMIT, we extract both the content factor ($s$, corresponds to the speaker identity) and the dynamic factors ($z_{1:T}$, correspond to the linguistic content) for the sequences. And for each of them, we compute the cosine similarity of representations between pairs of sequences, based on which we perform thresholding to classify if the two sequences are produced by the same speaker. Through varying the threshold, we compute the EER for the speaker verification task \citep{hsu2017unsupervised, li2018disentangled}. 
For a well disentangled latent space, we expect the EER on $s$ (content EER) to be low, while the EER on $z_{1:T}$ (motion EER) to be high. 
The EER results are given in Table~\ref{tab:timit}. Compared with the previous SOTA by R-WAE, C-DSVAE reduces the content EER by over 10\% relatively. 

\begin{figure}[t]
  \begin{subfigure}{0.495\textwidth}
    \includegraphics[width=\textwidth]{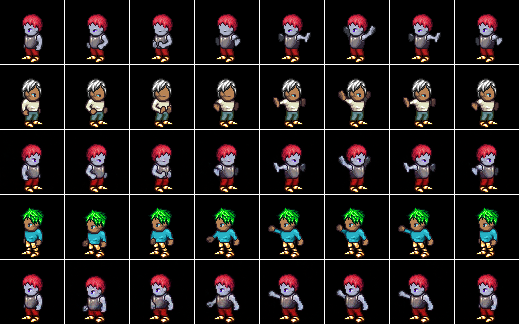}
  \end{subfigure}
  \hfill
  \begin{subfigure}{0.495\textwidth}
    \includegraphics[width=\textwidth]{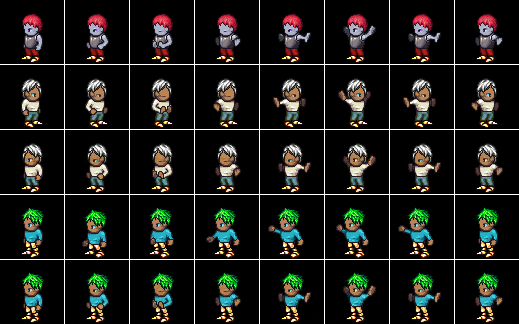}
  \end{subfigure}
  \caption{
  Manipulate the static and dynamic factors on Sprites. Row 1, 2, 4 are the raw test input sequences, while row 3, 5 are manipulated generations. \textbf{Left panel}: Row 3 uses $s$ from row 1 and $z_{1:T}$ from row 2. Row 5 uses $s$ from row 1 and $z_{1:T}$ from row 4. \textbf{Right panel}: Row 3 uses $z_{1:T}$ from row 1 and $s$ from row 2. Row 5 uses $z_{1:T}$ from row 1 and $s$ from row 4.}
  \label{fig:sprites}
\end{figure}

\begin{figure}[t]
  \begin{subfigure}{0.495\textwidth}
    \includegraphics[width=\textwidth]{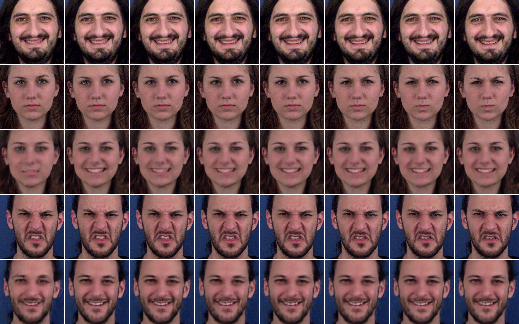}
    \caption{C-DSVAE}
  \end{subfigure}
  \hfill
  \begin{subfigure}{0.495\textwidth}
    \includegraphics[width=\textwidth]{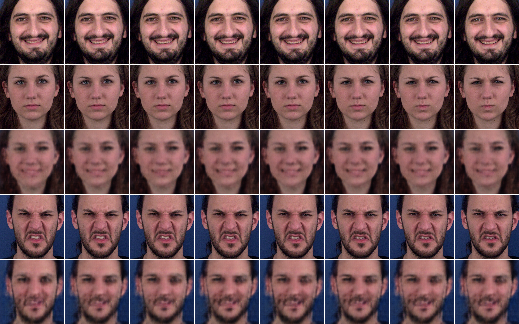}
    \caption{S3VAE}
  \end{subfigure}
  \caption{Fix the facial expression on MUG. Row 1, 2, 4 are the raw test sequences with different facial expressions. We replace $z_{1:T}$ of row 2, 4 with $z_{1:T}$ of row 1 and compare the performance of C-DSVAE and S3VAE. Hence row 3, 5 are expected to display happiness as well. Our C-DSVAE's generations are clean and sharp, while S3VAE produces blurred sequences (zoom in to see the details).}
  \label{fig:mug}
\end{figure}
\begin{figure}[h!]
  \begin{subfigure}{0.495\textwidth}
    \includegraphics[width=\textwidth]{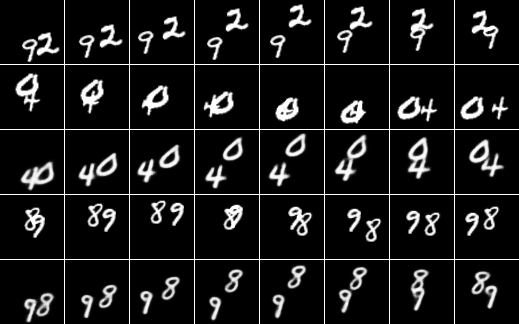}
    \caption{C-DSVAE}
    \label{fig:smmnist_cdsvae}
  \end{subfigure}
  \hfill
  \begin{subfigure}{0.495\textwidth}
    \includegraphics[width=\textwidth]{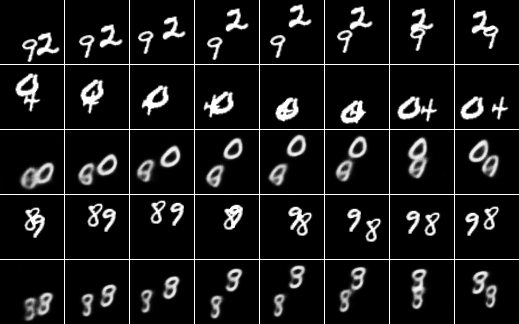}
    \caption{R-WAE}
    \label{fig:smmnist_rwae}
  \end{subfigure}
  \caption{Fix the movement and replace the digits. Row 1, 2, 4 are input test sequences with different digits and movements. Row 3, 5 use $z_{1:T}$ from row 1 while retaining their own $s$. The sequences generated by C-DSVAE are clean and consistent, while R-WAE sometimes produces blurred or incorrect digits.
  }
  \label{fig:smmnist}
\end{figure}
\begin{figure}[h!]
\centering
\begin{tabular}{@{}c@{\hspace{0.05\linewidth}}c@{}}
    \includegraphics[width=0.45\linewidth]{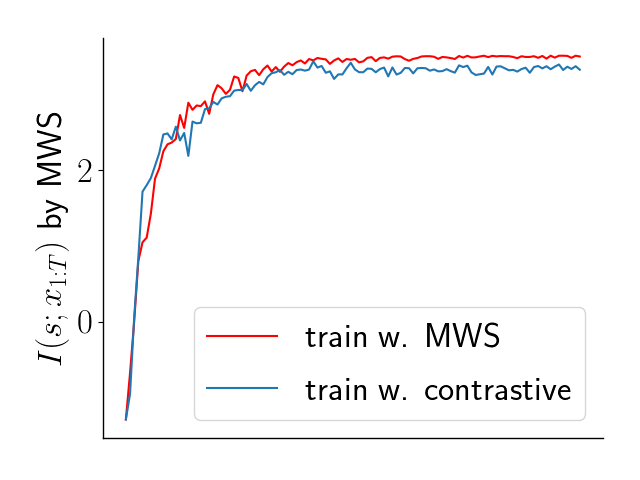}
&
    \includegraphics[width=0.45\linewidth]{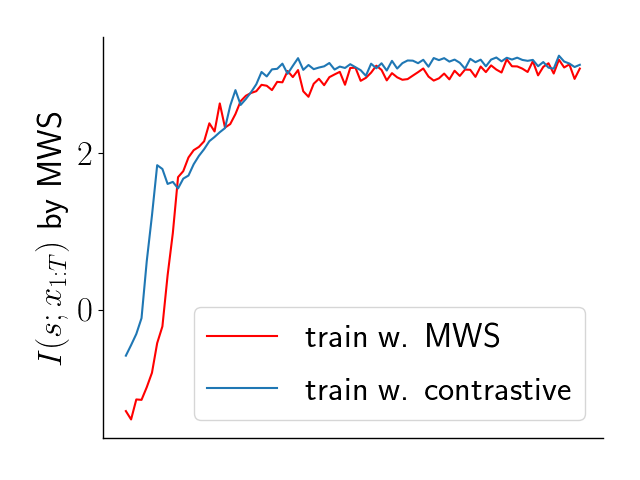} \\[-1ex]
    training time & training time 
  \end{tabular}
  \caption{
  Learning curves of the MI terms (estimated by MWS) between latent factors and the input sequences, when the training uses either MWS or the contrastive estimation for the MI terms. Though learning with different estimation methods, the MI curves for the two estimation methods are close to each other.
  \textbf{Left}: Sprites. \textbf{Right}: TIMIT. 
  }
  \label{fig:mi_sprites_fz_xs}
\end{figure}

\subsection{Qualitative Results}
\label{ss:qual}

We now compare different methods qualitatively with the following tasks: given the extracted static factor $s$ and dynamic factors $z_{1:T}$, we fix one of them and manipulate the other to see if we can observe the desired variations in the generated sequences. Figure~\ref{fig:sprites} gives example generations for Sprites. 
We observe that the character identities are well preserved when we fix $s$ (left) and the motion is well maintained when we fix $z_{1:T}$ (right), demonstrating the clean disentanglement by C-DSVAE.
Figure~\ref{fig:mug} compares the generated sequences when the motion (facial expression) is fixed for C-DSVAE and S3VAE on MUG. We replace $z_{1:T}$ for each input sequence with the one from the first row, and therefore the generated sequences are expected to have the smiling expression. C-DSVAE and S3VAE both generate recognizable smiling faces and preserve the individual's identity, but the sequences generated by S3VAE have blurs and are less sharp. In Figure~\ref{fig:smmnist}, we fix the digit movements and take the content (digit identities) from various other sequences. Entangled factors might cause the extracted $z_{1:T}$ to carry information of $s$. Row 3 of Figure~\ref{fig:smmnist_rwae} shows such an example produced by R-WAE where the motion factor of one digit carries the content information of "9" from row 1. 

Note that S3VAE and R-WAE are both capable of generating recognizable videos, which is validated by our classifier $\mathcal C$ with good accuracies. Nevertheless, the results here show that C-DSVAE can learn better disentangled factors for the high-quality generation. 
Results of additional generation tasks can be found in Appendix~\ref{ss:additional_expts},
where we swap the latent factors from different sequences.

\subsection{Other analysis}
\label{ss:analysis}

As mentioned in Sec~\ref{s:method}, we attempt to estimate the MI terms using either MWS or the contrastive estimation with augmentations.
In Figure~\ref{fig:mi_sprites_fz_xs}, we show the MWS estimation of $I(s;x_{1:T})$ improves when the C-DSVAE training objective optimizes the contrastive estimation of $I(s;x_{1:T})$. This means the contrastive estimation could be a surrogate for MI in generative models, and its additional inductive bias can be a main contributor to the superior performance (see Appendix~\ref{s:ablation} for quantitative comparisons).


Regarding the augmentations, it is intuitive to adopt both the content augmentation and the motion augmentation. Having both of them can not only boost the learning of the dynamic or static factors, but also improves the disentanglement. Adding only one of them could lead to unbalanced or entangled representation learning. As shown in Table~\ref{tab:abl_mug}, content augmentation alone can improve the accuracy by 3.8\%, and motion augmentation alone can improve the accuracy by 7.1\%. Two augmentations together can bring a gain of 9.1\%. Similarly, in Table~\ref{tab:abl_timit}, though adding only content augmentation or motion augmentation can outperform DSVAE, combining them leads to the largest gain. 

Furthermore, in Appendix~\ref{ss:mi_sz}, we show that even if we don’t include $I_q(s,z_{1:T})$ in the optimization, training on just the contrastive terms with augmentations can already lead to the decrease of $I_q(s,z_{1:T})$. In Appendix~\ref{s:latent_traversal}, the transition of digits illustrates our learnt latent space is smooth and meaningful.

\begin{table}[t]
\centering
\caption{Ablation study of the augmentations on SMMNIST.}
\label{tab:abl_mug}
\begin{tabular}{ccccc}
\toprule
Methods & Acc$\uparrow$ & IS$\uparrow$ & H(y|x)$\downarrow$ & H(y)$\uparrow$ \\
\midrule
DSVAE & 88.19\% & 6.210 & 0.185 & 2.011 \\
C-DSVAE w/o content aug & 92.02\% & 6.360 & 0.184 & 2.083 \\
C-DSVAE w/o motion aug & 95.25\% & 6.471 & 0.175 & 2.093 \\
C-DSVAE & 97.84\% & 7.163 & 0.145 & 2.176 \\
\bottomrule
\end{tabular}
\end{table}
\begin{table}[t]
\centering
\caption{Ablation study of the augmentations on TIMIT.}
\label{tab:abl_timit}
\begin{tabular}{ccc}
\toprule
Methods &  content EER$\downarrow$ & motion EER$\uparrow$\\
\midrule
DSVAE & 5.64\% & 19.20\% \\
C-DSVAE w/o content aug & 5.09\% & 24.30\% \\
C-DSVAE w/o motion aug & 4.31\% & 31.09\% \\
C-DSVAE & 4.03\% & 31.81\% \\
\bottomrule
\end{tabular}
\end{table}


\section{Conclusion}
\label{s:concls}

In this paper, we propose Contrastively Disentangled Sequential Variational Autoencoder (C-DSVAE) to learn disentangled static and dynamic latent factors for sequence data without external supervision. 
Our learning objective is a novel ELBO derived differently from prior works, and naturally encourages disentanglement.
C-DSVAE uses contrastive estimations of the MI terms to further inject the inductive biases.
Our method achieves the state-of-the-art performance on multiple datasets, in terms of the disentanglment metrics and the generation quality. 
In the future, we plan to extend C-DSVAE to other domains, such as text, biology, agriculture and weather prediction. We will also improve our model's ability to capture long-range dependencies. 

\section{Acknowledgment}
This work is supported by Hatch Federal Capacity Funds "3110006036 Hatch Bobea NYC-121437".

\bibliographystyle{unsrtnat}
\bibliography{neurips21.bib}

\begin{thebibliography}{57}
\providecommand{\natexlab}[1]{#1}
\providecommand{\url}[1]{\texttt{#1}}
\expandafter\ifx\csname urlstyle\endcsname\relax
  \providecommand{\doi}[1]{doi: #1}\else
  \providecommand{\doi}{doi: \begingroup \urlstyle{rm}\Url}\fi

\bibitem[Bengio et~al.(2013)Bengio, Courville, and
  Vincent]{bengio2013representation}
Yoshua Bengio, Aaron Courville, and Pascal Vincent.
\newblock Representation learning: A review and new perspectives.
\newblock \emph{IEEE transactions on pattern analysis and machine
  intelligence}, 35\penalty0 (8):\penalty0 1798--1828, 2013.

\bibitem[Oord et~al.(2018)Oord, Li, and Vinyals]{oord2018representation}
Aaron van~den Oord, Yazhe Li, and Oriol Vinyals.
\newblock Representation learning with contrastive predictive coding.
\newblock \emph{arXiv preprint arXiv:1807.03748}, 2018.

\bibitem[Hjelm et~al.(2018)Hjelm, Fedorov, Lavoie-Marchildon, Grewal, Bachman,
  Trischler, and Bengio]{hjelm2018learning}
R~Devon Hjelm, Alex Fedorov, Samuel Lavoie-Marchildon, Karan Grewal, Phil
  Bachman, Adam Trischler, and Yoshua Bengio.
\newblock Learning deep representations by mutual information estimation and
  maximization.
\newblock In \emph{International Conference on Learning Representations}, 2018.

\bibitem[Chen et~al.(2020)Chen, Kornblith, Norouzi, and Hinton]{chen2020simple}
Ting Chen, Simon Kornblith, Mohammad Norouzi, and Geoffrey Hinton.
\newblock A simple framework for contrastive learning of visual
  representations.
\newblock In \emph{International Conference on Machine Learning}, pages
  1597--1607. PMLR, 2020.

\bibitem[Golinski et~al.(2020)Golinski, Pourreza, Yang, Sautiere, and
  Cohen]{golinski2020feedback}
Adam Golinski, Reza Pourreza, Yang Yang, Guillaume Sautiere, and Taco~S Cohen.
\newblock Feedback recurrent autoencoder for video compression.
\newblock In \emph{Proceedings of the Asian Conference on Computer Vision},
  2020.

\bibitem[Peters et~al.(2018)Peters, Neumann, Iyyer, Gardner, Clark, Lee, and
  Zettlemoyer]{peters2018deep}
Matthew~E Peters, Mark Neumann, Mohit Iyyer, Matt Gardner, Christopher Clark,
  Kenton Lee, and Luke Zettlemoyer.
\newblock Deep contextualized word representations.
\newblock In \emph{Proceedings of NAACL-HLT}, pages 2227--2237, 2018.

\bibitem[Devlin et~al.(2019)Devlin, Chang, Lee, and Toutanova]{devlin2018bert}
Jacob Devlin, Ming-Wei Chang, Kenton Lee, and Kristina Toutanova.
\newblock Bert: Pre-training of deep bidirectional transformers for language
  understanding.
\newblock In \emph{Proceedings of the 2019 Conference of the North American
  Chapter of the Association for Computational Linguistics: Human Language
  Technologies, Volume 1 (Long and Short Papers)}, pages 4171--4186, 2019.

\bibitem[Brown et~al.(2020)Brown, Mann, Ryder, Subbiah, Kaplan, Dhariwal,
  Neelakantan, Shyam, Sastry, Askell, Agarwal, Herbert-Voss, Krueger, Henighan,
  Child, Ramesh, Ziegler, Wu, Winter, Hesse, Chen, Sigler, Litwin, Gray, Chess,
  Clark, Berner, McCandlish, Radford, Sutskever, and Amodei]{brown2020language}
Tom Brown, Benjamin Mann, Nick Ryder, Melanie Subbiah, Jared~D Kaplan, Prafulla
  Dhariwal, Arvind Neelakantan, Pranav Shyam, Girish Sastry, Amanda Askell,
  Sandhini Agarwal, Ariel Herbert-Voss, Gretchen Krueger, Tom Henighan, Rewon
  Child, Aditya Ramesh, Daniel Ziegler, Jeffrey Wu, Clemens Winter, Chris
  Hesse, Mark Chen, Eric Sigler, Mateusz Litwin, Scott Gray, Benjamin Chess,
  Jack Clark, Christopher Berner, Sam McCandlish, Alec Radford, Ilya Sutskever,
  and Dario Amodei.
\newblock Language models are few-shot learners.
\newblock In \emph{Advances in Neural Information Processing Systems},
  volume~33, pages 1877--1901, 2020.

\bibitem[Baevski and Mohamed(2020)]{baevski2019effectiveness}
Alexei Baevski and Abdelrahman Mohamed.
\newblock Effectiveness of self-supervised pre-training for asr.
\newblock In \emph{IEEE International Conference on Acoustics, Speech and
  Signal Processing (ICASSP)}, pages 7694--7698. IEEE, 2020.

\bibitem[Chung and Glass(2020)]{chung2020generative}
Yu-An Chung and James Glass.
\newblock Generative pre-training for speech with autoregressive predictive
  coding.
\newblock In \emph{IEEE International Conference on Acoustics, Speech and
  Signal Processing (ICASSP)}, pages 3497--3501. IEEE, 2020.

\bibitem[Wang et~al.(2020)Wang, Tang, and Livescu]{wang2020unsupervised}
Weiran Wang, Qingming Tang, and Karen Livescu.
\newblock Unsupervised pre-training of bidirectional speech encoders via masked
  reconstruction.
\newblock In \emph{IEEE International Conference on Acoustics, Speech and
  Signal Processing (ICASSP)}, pages 6889--6893. IEEE, 2020.

\bibitem[Baevski et~al.(2020)Baevski, Zhou, Mohamed, and
  Auli]{baevski2020wav2vec}
Alexei Baevski, Yuhao Zhou, Abdelrahman Mohamed, and Michael Auli.
\newblock wav2vec 2.0: A framework for self-supervised learning of speech
  representations.
\newblock \emph{Advances in Neural Information Processing Systems}, 33, 2020.

\bibitem[Bai et~al.(2021)Bai, Wang, Zhou, and Xiong]{bai2021representation}
Junwen Bai, Weiran Wang, Yingbo Zhou, and Caiming Xiong.
\newblock Representation learning for sequence data with deep autoencoding
  predictive components.
\newblock In \emph{International Conference on Learning Representations}, 2021.

\bibitem[Villegas et~al.(2017)Villegas, Yang, Hong, Lin, and
  Lee]{villegas2017decomposing}
Ruben Villegas, Jimei Yang, Seunghoon Hong, Xunyu Lin, and Honglak Lee.
\newblock Decomposing motion and content for natural video sequence prediction.
\newblock In \emph{International Conference on Learning Representations}, 2017.

\bibitem[Denton and Birodkar(2017)]{denton2017unsupervised}
Emily~L Denton and vighnesh Birodkar.
\newblock Unsupervised learning of disentangled representations from video.
\newblock In I.~Guyon, U.~V. Luxburg, S.~Bengio, H.~Wallach, R.~Fergus,
  S.~Vishwanathan, and R.~Garnett, editors, \emph{Advances in Neural
  Information Processing Systems}, volume~30, 2017.

\bibitem[Zhu et~al.(2018)Zhu, Elhoseiny, Liu, Peng, and
  Elgammal]{zhu2018generative}
Yizhe Zhu, Mohamed Elhoseiny, Bingchen Liu, Xi~Peng, and Ahmed Elgammal.
\newblock A generative adversarial approach for zero-shot learning from noisy
  texts.
\newblock In \emph{Proceedings of the IEEE conference on computer vision and
  pattern recognition}, pages 1004--1013, 2018.

\bibitem[Tian et~al.(2020)Tian, Ren, Chai, Olszewski, Peng, Metaxas, and
  Tulyakov]{tian2021good}
Yu~Tian, Jian Ren, Menglei Chai, Kyle Olszewski, Xi~Peng, Dimitris~N Metaxas,
  and Sergey Tulyakov.
\newblock A good image generator is what you need for high-resolution video
  synthesis.
\newblock In \emph{International Conference on Learning Representations}, 2020.

\bibitem[Han et~al.(2021)Han, Min, Han, Li, and Zhang]{han2021disentangled}
Jun Han, Martin~Renqiang Min, Ligong Han, Li~Erran Li, and Xuan Zhang.
\newblock Disentangled recurrent wasserstein autoencoder.
\newblock In \emph{International Conference on Learning Representations}, 2021.

\bibitem[Chen et~al.(2016)Chen, Duan, Houthooft, Schulman, Sutskever, and
  Abbeel]{chen2016infogan}
Xi~Chen, Yan Duan, Rein Houthooft, John Schulman, Ilya Sutskever, and Pieter
  Abbeel.
\newblock Infogan: Interpretable representation learning by information
  maximizing generative adversarial nets.
\newblock In \emph{Advances in Neural Information Processing Systems},
  volume~29, pages 2172--2180, 2016.

\bibitem[Higgins et~al.(2016)Higgins, Matthey, Pal, Burgess, Glorot, Botvinick,
  Mohamed, and Lerchner]{higgins2016beta}
Irina Higgins, Loic Matthey, Arka Pal, Christopher Burgess, Xavier Glorot,
  Matthew Botvinick, Shakir Mohamed, and Alexander Lerchner.
\newblock beta-vae: Learning basic visual concepts with a constrained
  variational framework.
\newblock \emph{International Conference on Learning Representations}, 2016.

\bibitem[Fraccaro et~al.(2017)Fraccaro, Kamronn, Paquet, and
  Winther]{fraccaro2017disentangled}
Marco Fraccaro, Simon Kamronn, Ulrich Paquet, and Ole Winther.
\newblock A disentangled recognition and nonlinear dynamics model for
  unsupervised learning.
\newblock In I.~Guyon, U.~V. Luxburg, S.~Bengio, H.~Wallach, R.~Fergus,
  S.~Vishwanathan, and R.~Garnett, editors, \emph{Advances in Neural
  Information Processing Systems}, volume~30, 2017.

\bibitem[Kim and Mnih(2018)]{kim2018disentangling}
Hyunjik Kim and Andriy Mnih.
\newblock Disentangling by factorising.
\newblock In \emph{International Conference on Machine Learning}, pages
  2649--2658. PMLR, 2018.

\bibitem[Chen et~al.(2018)Chen, Li, Grosse, and Duvenaud]{chen2018isolating}
Ricky T.~Q. Chen, Xuechen Li, Roger~B Grosse, and David~K Duvenaud.
\newblock Isolating sources of disentanglement in variational autoencoders.
\newblock In \emph{Advances in Neural Information Processing Systems},
  volume~31, 2018.

\bibitem[Locatello et~al.(2019)Locatello, Bauer, Lucic, Raetsch, Gelly,
  Sch{\"o}lkopf, and Bachem]{locatello2019challenging}
Francesco Locatello, Stefan Bauer, Mario Lucic, Gunnar Raetsch, Sylvain Gelly,
  Bernhard Sch{\"o}lkopf, and Olivier Bachem.
\newblock Challenging common assumptions in the unsupervised learning of
  disentangled representations.
\newblock In \emph{International Conference on Machine Learning}, pages
  4114--4124. PMLR, 2019.

\bibitem[Khemakhem et~al.(2020)Khemakhem, Kingma, Monti, and
  Hyvarinen]{khemakhem2020variational}
Ilyes Khemakhem, Diederik Kingma, Ricardo Monti, and Aapo Hyvarinen.
\newblock Variational autoencoders and nonlinear ica: A unifying framework.
\newblock In \emph{International Conference on Artificial Intelligence and
  Statistics}, pages 2207--2217. PMLR, 2020.

\bibitem[Locatello et~al.(2020)Locatello, Poole, R{\"a}tsch, Sch{\"o}lkopf,
  Bachem, and Tschannen]{locatello20a}
Francesco Locatello, Ben Poole, Gunnar R{\"a}tsch, Bernhard Sch{\"o}lkopf,
  Olivier Bachem, and Michael Tschannen.
\newblock Weakly-supervised disentanglement without compromises.
\newblock In \emph{International Conference on Machine Learning}, pages
  6348--6359. PMLR, 2020.

\bibitem[Hochreiter and Schmidhuber(1997)]{hochreiter1997long}
Sepp Hochreiter and J{\"u}rgen Schmidhuber.
\newblock Long short-term memory.
\newblock \emph{Neural computation}, 9\penalty0 (8):\penalty0 1735--1780, 1997.

\bibitem[Kingma and Welling(2014)]{kingma2014auto}
Diederik~P Kingma and Max Welling.
\newblock Auto-encoding variational bayes.
\newblock In \emph{International Conference on Learning Representations}, 2014.

\bibitem[Chung et~al.(2015)Chung, Kastner, Dinh, Goel, Courville, and
  Bengio]{chung2015recurrent}
Junyoung Chung, Kyle Kastner, Laurent Dinh, Kratarth Goel, Aaron~C Courville,
  and Yoshua Bengio.
\newblock A recurrent latent variable model for sequential data.
\newblock In C.~Cortes, N.~Lawrence, D.~Lee, M.~Sugiyama, and R.~Garnett,
  editors, \emph{Advances in Neural Information Processing Systems}, volume~28,
  2015.

\bibitem[Goyal et~al.(2017)Goyal, Sordoni, C{\^o}t{\'e}, Ke, and
  Bengio]{goyal2017z}
Anirudh Goyal, Alessandro Sordoni, Marc-Alexandre C{\^o}t{\'e}, Nan~Rosemary
  Ke, and Yoshua Bengio.
\newblock Z-forcing: Training stochastic recurrent networks.
\newblock In \emph{Advances in Neural Information Processing Systems},
  volume~30, 2017.

\bibitem[Krishnan et~al.(2015)Krishnan, Shalit, and Sontag]{krishnan2015deep}
Rahul~G Krishnan, Uri Shalit, and David Sontag.
\newblock Deep kalman filters.
\newblock \emph{arXiv preprint arXiv:1511.05121}, 2015.

\bibitem[Li and Mandt(2018)]{li2018disentangled}
Yingzhen Li and Stephan Mandt.
\newblock Disentangled sequential autoencoder.
\newblock In \emph{International Conference on Machine Learning}, pages
  5670--5679. PMLR, 2018.

\bibitem[Zhu et~al.(2020)Zhu, Min, Kadav, and Graf]{zhu2020s3vae}
Yizhe Zhu, Martin~Renqiang Min, Asim Kadav, and Hans~Peter Graf.
\newblock S3vae: Self-supervised sequential vae for representation
  disentanglement and data generation.
\newblock In \emph{Proceedings of the IEEE/CVF Conference on Computer Vision
  and Pattern Recognition}, pages 6538--6547, 2020.

\bibitem[Bowman et~al.(2016)Bowman, Vilnis, Vinyals, Dai, Jozefowicz, and
  Bengio]{bowman2015generating}
Samuel Bowman, Luke Vilnis, Oriol Vinyals, Andrew Dai, Rafal Jozefowicz, and
  Samy Bengio.
\newblock Generating sentences from a continuous space.
\newblock In \emph{Proceedings of The 20th SIGNLL Conference on Computational
  Natural Language Learning}, pages 10--21, 2016.

\bibitem[Tomczak and Welling(2018)]{tomczak2018vae}
Jakub Tomczak and Max Welling.
\newblock Vae with a vampprior.
\newblock In \emph{International Conference on Artificial Intelligence and
  Statistics}, 2018.

\bibitem[Alemi et~al.(2018)Alemi, Poole, Fischer, Dillon, Saurous, and
  Murphy]{alemi2018fixing}
Alexander Alemi, Ben Poole, Ian Fischer, Joshua Dillon, Rif~A Saurous, and
  Kevin Murphy.
\newblock Fixing a broken elbo.
\newblock In \emph{International Conference on Learning Representations}, 2018.

\bibitem[Takahashi et~al.(2019)Takahashi, Iwata, Yamanaka, Yamada, and
  Yagi]{takahashi2019variational}
Hiroshi Takahashi, Tomoharu Iwata, Yuki Yamanaka, Masanori Yamada, and Satoshi
  Yagi.
\newblock Variational autoencoder with implicit optimal priors.
\newblock In \emph{Proceedings of the AAAI Conference on Artificial
  Intelligence}, pages 5066--5073, 2019.

\bibitem[Tschannen et~al.(2018)Tschannen, Bachem, and
  Lucic]{tschannen2018recent}
Michael Tschannen, Olivier Bachem, and Mario Lucic.
\newblock Recent advances in autoencoder-based representation learning.
\newblock \emph{arXiv preprint arXiv:1812.05069}, 2018.

\bibitem[Akuzawa et~al.(2021)Akuzawa, Iwasawa, and
  Matsuo]{akuzawa2021information}
Kei Akuzawa, Yusuke Iwasawa, and Yutaka Matsuo.
\newblock Information-theoretic regularization for learning global features by
  sequential vae.
\newblock \emph{Machine Learning}, 110\penalty0 (8):\penalty0 2239--2266, 2021.

\bibitem[Bachman et~al.(2019)Bachman, Hjelm, and
  Buchwalter]{bachman2019learning}
Philip Bachman, R~Devon Hjelm, and William Buchwalter.
\newblock Learning representations by maximizing mutual information across
  views.
\newblock In \emph{Advances in Neural Information Processing Systems},
  volume~32, 2019.

\bibitem[Verhelst and Roelands(1993)]{verhelst1993overlap}
Werner Verhelst and Marc Roelands.
\newblock An overlap-add technique based on waveform similarity (wsola) for
  high quality time-scale modification of speech.
\newblock In \emph{IEEE International Conference on Acoustics, Speech, and
  Signal Processing}, volume~2, pages 554--557. IEEE, 1993.

\bibitem[Driedger and M{\"u}ller(2016)]{driedger2016review}
Jonathan Driedger and Meinard M{\"u}ller.
\newblock A review of time-scale modification of music signals.
\newblock \emph{Applied Sciences}, 6\penalty0 (2):\penalty0 57, 2016.

\bibitem[Khosla et~al.(2020)Khosla, Teterwak, Wang, Sarna, Tian, Isola,
  Maschinot, Liu, and Krishnan]{khosla2020supervised}
Prannay Khosla, Piotr Teterwak, Chen Wang, Aaron Sarna, Yonglong Tian, Phillip
  Isola, Aaron Maschinot, Ce~Liu, and Dilip Krishnan.
\newblock Supervised contrastive learning.
\newblock In \emph{Advances in Neural Information Processing Systems},
  volume~33, 2020.

\bibitem[Goodfellow et~al.(2014)Goodfellow, Pouget-Abadie, Mirza, Xu,
  Warde-Farley, Ozair, Courville, and Bengio]{NIPS2014_Ian}
Ian Goodfellow, Jean Pouget-Abadie, Mehdi Mirza, Bing Xu, David Warde-Farley,
  Sherjil Ozair, Aaron Courville, and Yoshua Bengio.
\newblock Generative adversarial nets.
\newblock In \emph{Advances in Neural Information Processing Systems},
  volume~27, 2014.

\bibitem[Tulyakov et~al.(2018)Tulyakov, Liu, Yang, and
  Kautz]{tulyakov2018mocogan}
Sergey Tulyakov, Ming-Yu Liu, Xiaodong Yang, and Jan Kautz.
\newblock Mocogan: Decomposing motion and content for video generation.
\newblock In \emph{2018 IEEE/CVF Conference on Computer Vision and Pattern
  Recognition (CVPR)}, pages 1526--1535. IEEE Computer Society, 2018.

\bibitem[Hsu et~al.(2017)Hsu, Zhang, and Glass]{hsu2017unsupervised}
Wei-Ning Hsu, Yu~Zhang, and James Glass.
\newblock Unsupervised learning of disentangled and interpretable
  representations from sequential data.
\newblock In \emph{Advances in Neural Information Processing Systems},
  volume~30, 2017.

\bibitem[Cheng et~al.(2020)Cheng, Min, Shen, Malon, Zhang, Li, and
  Carin]{cheng2020improving}
Pengyu Cheng, Martin~Renqiang Min, Dinghan Shen, Christopher Malon, Yizhe
  Zhang, Yitong Li, and Lawrence Carin.
\newblock Improving disentangled text representation learning with
  information-theoretic guidance.
\newblock In \emph{Proceedings of the 58th Annual Meeting of the Association
  for Computational Linguistics}, pages 7530--7541, 2020.

\bibitem[Hyvarinen and Morioka(2016)]{NIPS2016_d305281f}
Aapo Hyvarinen and Hiroshi Morioka.
\newblock Unsupervised feature extraction by time-contrastive learning and
  nonlinear ica.
\newblock In D.~Lee, M.~Sugiyama, U.~Luxburg, I.~Guyon, and R.~Garnett,
  editors, \emph{Advances in Neural Information Processing Systems}, volume~29,
  2016.

\bibitem[Hyvarinen and Morioka(2017)]{hyvarinen17a}
Aapo Hyvarinen and Hiroshi Morioka.
\newblock Nonlinear ica of temporally dependent stationary sources.
\newblock In \emph{Artificial Intelligence and Statistics}, pages 460--469.
  PMLR, 2017.

\bibitem[Gresele et~al.(2020)Gresele, Rubenstein, Mehrjou, Locatello, and
  Sch{\"o}lkopf]{gresele2019incomplete}
Luigi Gresele, Paul~K Rubenstein, Arash Mehrjou, Francesco Locatello, and
  Bernhard Sch{\"o}lkopf.
\newblock The incomplete rosetta stone problem: Identifiability results for
  multi-view nonlinear ica.
\newblock In \emph{Uncertainty in Artificial Intelligence}, pages 217--227.
  PMLR, 2020.

\bibitem[Reed et~al.(2015)Reed, Zhang, Zhang, and Lee]{reed2015deep}
Scott~E Reed, Yi~Zhang, Yuting Zhang, and Honglak Lee.
\newblock Deep visual analogy-making.
\newblock In C.~Cortes, N.~Lawrence, D.~Lee, M.~Sugiyama, and R.~Garnett,
  editors, \emph{Advances in Neural Information Processing Systems}, volume~28,
  2015.

\bibitem[Aifanti et~al.(2010)Aifanti, Papachristou, and
  Delopoulos]{aifanti2010mug}
Niki Aifanti, Christos Papachristou, and Anastasios Delopoulos.
\newblock The mug facial expression database.
\newblock In \emph{11th International Workshop on Image Analysis for Multimedia
  Interactive Services WIAMIS 10}, pages 1--4. IEEE, 2010.

\bibitem[Denton and Fergus(2018)]{denton2018stochastic}
Emily Denton and Rob Fergus.
\newblock Stochastic video generation with a learned prior.
\newblock In \emph{International Conference on Machine Learning}, pages
  1174--1183. PMLR, 2018.

\bibitem[Garofolo et~al.(1992)Garofolo, Lamel, Fisher, Fiscus, Pallett,
  Dahlgren, and Zue]{timit1992}
J.~Garofolo, Lori Lamel, W.~Fisher, Jonathan Fiscus, D.~Pallett, N.~Dahlgren,
  and V.~Zue.
\newblock Timit acoustic-phonetic continuous speech corpus.
\newblock \emph{Linguistic Data Consortium}, 1992.

\bibitem[Zhou et~al.(2017)Zhou, Zhang, and Wang]{zhou2017inception}
Zhiming Zhou, Weinan Zhang, and Jun Wang.
\newblock Inception score, label smoothing, gradient vanishing and log(d(x))
  alternative.
\newblock \emph{arXiv preprint arXiv:1708.01729}, 2017.

\bibitem[Dehak et~al.(2010)Dehak, Kenny, Dehak, Dumouchel, and
  Ouellet]{dehak2010front}
Najim Dehak, Patrick~J Kenny, R{\'e}da Dehak, Pierre Dumouchel, and Pierre
  Ouellet.
\newblock Front-end factor analysis for speaker verification.
\newblock \emph{IEEE Transactions on Audio, Speech, and Language Processing},
  19\penalty0 (4):\penalty0 788--798, 2010.

\bibitem[Kingma and Ba(2015)]{kingma2014adam}
Diederik~P Kingma and Jimmy Ba.
\newblock Adam: A method for stochastic optimization.
\newblock In \emph{International Conference on Learning Representations}, 2015.

\end{thebibliography}

\clearpage
\appendix
\section{Proofs}
\label{s:proofs}

\subsection{Derivation of the per-sequence ELBO}
\label{ss:derivation_of_per_seq_elbo}

For the input sequence $x_{1:T}$, we show the ELBO derived from the approximate posterior is a lower estimate of its log-likelihood. This proof below is adapted from the standard VAE framework \citep{kingma2014auto, zhu2020s3vae}, noticing that either the prior or the approximate posterior factorizes over $s$ and $z_{1:T}$.
\begin{equation*}
\begin{aligned}
& \log p(x_{1:T}) \\
\ge& -KL[q(s,z_{1:T}|x_{1:T})||p(s,z_{1:T}|x_{1:T})]+\log p(x_{1:T})\\
=&\mathbb E_{q(s, z_{1:T}|x_{1:T})} \left[ \log p(s,z_{1:T}|x_{1:T}) - \log q(s,z_{1:T}|x_{1:T}) + \log p(x_{1:T}) \right] \\
=&\mathbb E_{q(s,z_{1:T}|x_{1:T})}[\log p(x_{1:T}|s, z_{1:T})-\log q(s,z_{1:T}|x_{1:T})+\log p(s,z_{1:T})]\\
=&\mathbb E_{q(s,z_{1:T}|x_{1:T})}[\log p(x_{1:T}|s, z_{1:T})-\log q(s|x_{1:T}) - \log p(z_{1:T}|x_{1:T})+\log p(s) + \log p(z_{1:T})]\\
=&\mathbb E_{q(z_{1:T}, s|x_{1:T})} \left[\log p(x_{1:T}|s, z_{1:T})]-KL[q(s|x_{1:T})||p(s)]-KL[q(z_{1:T}|x_{1:T})||p(z_{1:T})\right].
\end{aligned}  
\end{equation*}

\subsection{Proof of Theorem 1}
\label{ss:thm1_proof}

Let $p_D$ be the empirical data distribution, assigning probability mass $1/N$ for each of the $N$ training sequences $D$. Define the aggregated posteriors as follows:
\begin{align*} \label{eq:aggregated-s}
q(s) & = \mathbb E_{x_{1:T}\sim p_D} [q(s|x_{1:T})] = \frac{1}{N} \sum_{x_{1:T}\in D} q(s|x_{1:T}),
\\
~\\
q(z_{1:T}) & = \mathbb E_{x_{1:T}\sim p_D} [q(z_{1:T}|x_{1:T})] = \frac{1}{N} \sum_{x_{1:T}\in D} q(z_{1:T}|x_{1:T}),
\\
~\\
q(s,z_{1:T}) & = \mathbb E_{x_{1:T}\sim p_D} [q(s|x_{1:T}) q(z_{1:T}|x_{1:T})] = \frac{1}{N} \sum_{x_{1:T}\in D} q(s|x_{1:T}) q(z_{1:T}|x_{1:T}).
\end{align*}

With these definitions, we have
\begin{equation}
\begin{aligned}
 &\mathbb E_{x_{1:T}\sim p_D}[KL[q(s|x_{1:T})||p(s)]] \\
=&\mathbb E_{x_{1:T}\sim p_D}\mathbb E_{q(s|x_{1:T})}[\log q(s|x_{1:T}) - \log q(s) + \log q(s) - \log p(s)] \\
=&\mathbb E_{q(x_{1:T}, s)} \log \left[ \frac{q(s|x_{1:T})}{q(s)} \right]  + \mathbb E_{q(x_{1:T}, s)} [\log q(s)-\log p(s)] \\
=& I_q (x_{1:T};s) + KL [q(s)||p(s)].
\end{aligned}
\end{equation}
In other words, 
\begin{equation}
\begin{aligned}
KL [q(s)||p(s)]=\mathbb E_{x_{1:T}\sim p_D}[KL[q(s|x_{1:T})||p(s)]]-I_q(x_{1:T};s).
\end{aligned}
\label{eq:kl_f}
\end{equation}

Similarly, we have 
\begin{gather} \label{eq:kl_z}
 KL[q(z_{1:T})||p(z_{1:T})]=E_{x_{1:T}\sim p_D}[KL[q(z_{1:T}|x_{1:T})||p(z_{1:T})]] - I_q(x_{1:T};z_{1:T}).
\end{gather}

We are now ready to prove the theorem.
We derive a dataset ELBO by subtracting a different KL-divergence from the data log-likelihood:
\begin{equation}
\begin{aligned} 
&\frac{1}{N} \sum_{x_{1:T}\in D} \log p(x_{1:T}) =\mathbb E_{x_{1:T}\sim p_D}[\log p(x_{1:T})]\\
\ge& \mathbb E_{x_{1:T}\sim p_D}[\log p(x_{1:T})-KL[q(s, z_{1:T})||p(s, z_{1:T}|x_{1:T})]] \\
=& \mathbb E_{x_{1:T}\sim p_D}[\mathbb E_{q(s, z_{1:T}|x_{1:T})}[\log p(x_{1:T})-(\log q(s, z_{1:T})-\log p(s, z_{1:T}|x_{1:T}))]] \\
=& \mathbb E_{x_{1:T}\sim p_D}[\mathbb E_{q(s, z_{1:T}|x_{1:T})}[\log p(x_{1:T})-\log q(s, z_{1:T})+\log p(s, z_{1:T}|x_{1:T})]] \\
=& \mathbb E_{x_{1:T}\sim p_D}[\mathbb E_{q(s, z_{1:T}|x_{1:T})}[\log p(x_{1:T})-\log q(s, z_{1:T})\\
&\hspace{9em} +\log p(x_{1:T}|s, z_{1:T})+\log p(s, z_{1:T})-\log p(x_{1:T})]]\\
=& \mathbb E_{x_{1:T}\sim p_D}[\mathbb E_{q(s, z_{1:T}|x_{1:T})}[\log p(x_{1:T}|s, z_{1:T})-\log q(s, z_{1:T})+\log p(s, z_{1:T})]]\\
=& \mathbb E_{x_{1:T}\sim p_D}[\mathbb E_{q(s, z_{1:T}|x_{1:T})}[\log p(x_{1:T}|s, z_{1:T})]] - \mathbf{KL[q(s,z_{1:T})||p(s,z_{1:T})}]\\
=& \mathbb E_{x_{1:T}\sim p_D}[\mathbb E_{q(s, z_{1:T}|x_{1:T})}[\log p(x_{1:T}|s, z_{1:T})]]\\
&\hspace{5em} - \mathbb E_{x_{1:T}\sim p_D}[\mathbb E_{q(s, z_{1:T}|x_{1:T})}[\log q(s, z_{1:T})-\log p(s, z_{1:T})]]\\
=& \mathbb E_{x_{1:T}\sim p_D}[\mathbb E_{q(s, z_{1:T}|x_{1:T})}[\log p(x_{1:T}|s, z_{1:T})]]\\
& -\mathbb E_{x_{1:T}\sim p_D}[\mathbb E_{q(s, z_{1:T}|x_{1:T})}[\log q(s, z_{1:T})-\log q(s)q(z_{1:T})+\log q(s)q(z_{1:T})-\log p(s, z_{1:T})]]\\
=& \mathbb E_{x_{1:T}\sim p_D}[\mathbb E_{q(s, z_{1:T}|x_{1:T})}[\log p(x_{1:T}|s, z_{1:T})]]\\
&\hspace{5em} -\mathbb E_{x_{1:T}\sim p_D} \left[\mathbb E_{q(s, z_{1:T}|x_{1:T})}\left[\log \frac{q(s, z_{1:T})}{q(s)q(z_{1:T})}+\log \frac{q(s)q(z_{1:T})}{p(s, z_{1:T})}\right]\right]\\
=& \mathbb E_{x_{1:T}\sim p_D}[\mathbb E_{q(s, z_{1:T}|x_{1:T})}[\log p(x_{1:T}|s, z_{1:T})]]\\
&\hspace{5em} -I_q (s;z_{1:T})-\mathbb E_{x_{1:T}\sim p_D} \left[\mathbb E_{q(s, z_{1:T}|x_{1:T})} \left[\log \frac{q(s)q(z_{1:T})}{p(s)p(z_{1:T})}\right]\right]\\
=& \mathbb E_{x_{1:T}\sim p_D}[\mathbb E_{q(s, z_{1:T}|x_{1:T})}[\log p(x_{1:T}|s, z_{1:T})]]-I_q (s;z_{1:T})\\
&-\mathbb E_{x_{1:T}\sim p_D} \left[\mathbb E_{q(s, z_{1:T}|x_{1:T})}\left[\log \frac{q(s)}{p(s)}\right]\right]-\mathbb E_{x_{1:T}\sim p_D}\left[\mathbb E_{q(s, z_{1:T}|x_{1:T})}\left[\log \frac{q(z_{1:T})}{p(z_{1:T})}\right]\right]\\
=& \mathbb E_{x_{1:T}\sim p_D}[\mathbb E_{q(s, z_{1:T}|x_{1:T})}[\log p(x_{1:T}|s, z_{1:T})]]-I_q(s;z_{1:T})-KL[q(s)||p(s)] -KL[q(z_{1:T})||p(z_{1:T})]\\
=& \mathbb E_{x_{1:T}\sim p_D}[\mathbb E_{q(s, z_{1:T}|x_{1:T})}[\log p(x_{1:T}|s, z_{1:T})]]-I_q(s;z_{1:T})\\
&\hspace{5em} -(\mathbb E_{x_{1:T}\sim p_D}[KL[q(s|x_{1:T})||p(s)]]-I_q(s;x_{1:T}))\\
&\hspace{5em} -(\mathbb E_{x_{1:T}\sim p_D}[KL[q(z_{1:T}|x_{1:T})||p(z_{1:T})]]-I_q(z_{1:T};x_{1:T}))\\
=&
\mathbb E_{x_{1:T}\sim p_D} [ \mathbb E_{q(z_{1:T}, s|x_{1:T})}[\log p(x_{1:T}|s, z_{1:T})] \\
&\hspace{5em} - \mathbb E_{x_{1:T}\sim p_D} [KL[q(s|x_{1:T})||p(s)]] 
- \mathbb E_{x_{1:T}\sim p_D} [KL[q(z_{1:T}|x_{1:T})||p(z_{1:T})]]\\
&\hspace{5em} +I_q(s;x_{1:T})+I_q(z_{1:T};x_{1:T})-I_q(s;z_{1:T}).
\label{eq:elbo}
\end{aligned}
\end{equation}
where we have plugged in~\eqref{eq:kl_f} and \eqref{eq:kl_z} in the third to last step. The last equation of~\eqref{eq:elbo} is the dataset ELBO objective in the main text.

\subsection{MWS estimation of $I(s;z_{1:T})$}
\label{ss:mws_mi}

We estimate $I(s;z_{1:T})$ using minibatch weighted sampling (MWS). The estimation is simply adapted from \citep{chen2018isolating,zhu2020s3vae}:
\begin{equation}
\begin{aligned}
&I(s;z_{1:T}) = H(s) + H(z_{1:T}) - H(s,z_{1:T})\\
&H(s) = - \mathbb E_{q(s,z_{1:T})} [\log q(s)]\approx -\frac{1}{M}\sum_{i=1}^M \left[\log \sum_{j=1}^M q(s(x^i_{1:T})|x^j_{1:T})-\log(NM)\right]
\end{aligned}
\end{equation}
where $N$ is the dataset size and $M$ is the minibatch size. $H(z_{1:T}), H(s,z_{1:T})$ are estimated similarly, with the trajectory of $z_{1:T}$ sampled for each sequence in the minibatch.

\subsection{Contrastive estimation of MI}
\label{ss:contrast_est_mi}

Suppose $z_{1:T}$ is derived from the anchor sequence $x_{1:T}$, and we have one "positive" sequence $x^+_{1:T}$ as well as $n$ "negative" sequences $\{x^j_{1:T}\}_{j=1}^n$ in total. We have
\begin{equation}
\begin{aligned}
&\mathbb E_{p_D} \left[\log \frac{\frac{q(x^+_{1:T}|z_{1:T})}{q(x^+_{1:T})}}{\frac{q(x^+_{1:T}|z_{1:T})}{q(x^+_{1:T})} + \sum_{j=1}^n \frac{q(x^j_{1:T}|z_{1:T})}{q(x^j_{1:T})} } \right]\\
=&-\mathbb E_{p_D} \left[\log (1+\frac{q(x^+_{1:T})}{q(x^+_{1:T}|z_{1:T})}\sum_{j=1}^n \frac{q(x^j_{1:T}|z_{1:T})}{q(x^j_{1:T})}) \right]\\
\approx&-\mathbb E_{p_D} \left[\log \left(1+\frac{q(x^+_{1:T})}{q(x^+_{1:T}|z_{1:T})}\cdot n\mathbb E_{x^j_{1:T}} \left[\frac{q(x^j_{1:T}|z_{1:T})}{q(x^j_{1:T})}\right]\right)\right]\\
=&-\mathbb E_{p_D} \left[\log \left(1+\frac{q(x^+_{1:T})}{q(x^+_{1:T}|z_{1:T})}\cdot n \right)\right]\\
=&-\mathbb E_{p_D}\left[\log \left(\frac{1}{1+n}+\frac{q(x^+_{1:T})}{q(x^+_{1:T}|z_{1:T})}\cdot \frac{n}{1+n}\right)\right]-\log(n+1)\\
\le&-\mathbb E_{p_D} \left[\frac{1}{n+1}\log 1+\frac{n}{n+1}\log\frac{q(x^+_{1:T})}{q(x^+_{1:T}|z_{1:T})}\right]-\log(n+1)\\
=&\mathbb -\frac{n}{n+1}E_{p_D}\left[\log\frac{q(x^+_{1:T})}{q(x^+_{1:T}|z_{1:T})}\right]-\log(n+1)\\
=& \frac{n}{n+1} \mathbb E_{p_D} \left[\log\frac{q(x^+_{1:T}|z_{1:T})}{q(x^+_{1:T})}\right]-\log(n+1)\\
\approx& \frac{n}{n+1} I(x_{1:T};z_{1:T})-\log(n+1)
\end{aligned}
\end{equation}
where the first $\le$ step uses Jensen's inequality,
and the approximations by sampling follow the development of CPC~\citep{oord2018representation}. Similar derivations can also be obtained for $s$.


\paragraph{Use augmentation in contrastive estimation} Note that in CPC, for each frame, the positive example is a nearby frame. That is, CPC uses the temporal smoothness as the inductive bias for learning per-frame representations. In our setup however, we treat the entire trajectory $z_{1:T}$ as samples in contrastive estimation, and we must resort to other inductive biases for finding positive examples.

Here we provide another motivation for the use of augmented sequences as positive examples. Imagine that the latent factors are discrete, and the mapping from the latent factors $(s,z_{1:T})$ to observations $x_{1:T}$ is done by a deterministic mapping (while the factors remain random variables with prior distributions). Furthermore, the mapping is invertible in the sense that we could identify a unique $(s, z_{1:T})$ that generates $x_{1:T}$.
Then for two sequences $x_{1:T}^1$ and $x_{1:T}^2$ generated with common dynamic factors $z_{1:T}^*$ but different static factors $s^1$ and $s^2$, we have
\begin{gather*}
p(x_{1:T}^1) = p(z_{1:T}=z_{1:T}^*,s=s^1) = p(z_{1:T}=z_{1:T}^*)\, p(s=s^1),\\
p(x_{1:T}^1|z_{1:T}) = 1(z_{1:T}=z_{1:T}^*)\, p(s=s^1), \\
\frac{p(x_{1:T}^1|z_{1:T})}{p(x_{1:T})} = \frac{1(z_{1:T}=z_{1:T}^*)}{p(z_{1:T}=z_{1:T}^*)} = \frac{p(x_{1:T}^2|z_{1:T})}{p(x_{1:T})}.
\end{gather*}
The last equation is obtained by dividing the second line by the first line, where the $p(s=s^1)$ is conveniently canceled out. 
In other words, under the above simplifying assumptions, the probability ratio used in contrastive estimation is the same for the original sequence and the augmented sequence.

\subsection{Summary of objective function and estimations}

Though the objective function~\eqref{eq:elbo} has been proven to be an ELBO, we don't have to stick to the natural coefficients. For example, to control the information bottleneck, one can add coefficients to the KL terms. To improve the disentanglement, one can add coefficients to the MI terms. This leads to the final training objective~\eqref{eq:dsvae_mi_final}.

In practice, $\gamma=1$ gives good results and we mainly tune $\alpha, \beta$. As mentioned in Sec \ref{ss:augmentation}, $I(s;x_{1:T}), I(z_{1:T};x_{1:T})$ are estimated contrastively. $I(s;z_{1:T})$ is estimated through mini-batch weighted sampling (MWS) as discussed in~\ref{ss:mws_mi}. 
Note that we use contrastive learning here to maintain some invariance (static or dynamic factors) across different views, besides just estimating the MI terms. For tasks where no such invariance exists, MWS is still a good option for the estimation (e.g. $I(z_{1:T};s)$). More comparisons can be found in \ref{s:ablation}.

\section{Disentanglement Metrics}
\textbf{Accuracy} (Acc) measures how well the fixed factor can be identified by the classifier $C$. Given an input sequence $x_{1:T}$, the encoder would produce $z_{1:T}$ and $s$. If we randomly sample from the prior instead of the posterior of $s$ for decoding/generation, the classifier $C$ should still recognize the same motion captured by $z_{1:T}$ while identifying different contents induced by the random $s$. For example, in the Sprites dataset, if we randomly sample $s$ from the prior and fix $z_{1:T}$, we should see the generated characters with different colors performing the same action in the same direction. 

\textbf{Inception Score} ($IS$) computes the KL-divergence between the conditional predicted label distribution $p(y|x_{1:T})$ and marginal predicted label distribution $p(y)$ from $C$, $IS=\exp(\mathbb E_{p(x)}[KL[p(y|x)||p(y)]])$. $y$ is the predicted attribute such as color, action, expression, etc. $p(y|x_{1:T})$ is usually taken from the logits after the softmax layer. $p(y)$ is the marginalization across all the test samples, $p(y)=\frac{1}{N}\sum_{i=1}^N p(y|x)$. Since we hope the generated samples to be as diverse as possible, $IS$ is expected to be high.

\textbf{Inter-Entropy} is similar to $IS$ but measures the diversity solely on the marginal predicted label distribution $p(y)$,
$H(y) = -\sum_y p(y)\log p(y)$.
The higher $H(y)$ is, the more diverse generated sequences would be.

\textbf{Intra-Entropy} measures the entropy over $p(y|x)$, $H(y|x)=-\sum_y p(y|x)\log p(y|x)$. Lower intra-entropy means more confident predictions.

\textbf{Equal Error Rate} (EER) is only used in the TIMIT experiments for audio evaluation. It means the common value when the false rejection rate is equal to the false acceptance rate.

\section{Training Details}
\label{ap:training}

\subsection{Dataset}
\textbf{Sprites} is a cartoon character video dataset. Each character's (dynamic) motion is captured by three actions (walking, spellcasting, slashing) and three directions (left, front, right). The (static) content comprises of each character's skin color, tops color, pants color and hair color. Each color has six variants. Each sequence is composed of 8 frames of RGB images with size $64\times 64$. There are in total 1296 characters. 1000 of them are used for training and validation. The rest 296 characters are used for testing. 

\textbf{MUG} is a facial expression video dataset. The static factor corresponds to the person's identity. Each individual performs six expressions (motion): anger, fear, disgust, happiness, sadness and surprise. Each sequence contains 15 frames of RGB images with size $64\times 64$ (after resizing). There are in total 3528 videos with 52 people. Each video contains 50 to 160 frames. Length-15 clips are sampled from the raw videos. Each frame from the raw videos is reshaped to $64\times64$.

\textbf{SM-MNIST} (Stochastic Moving MNIST) records the random movements of two digits. Each sequence contains 15 gray scale images of size $64\times 64$. Individual digits are collected from MNIST. We adopt the S3VAE setup with two moving digits instead of just one.


\textbf{TIMIT} is a corpus of read speech for acoustic-phonetic studies and speech recognition. The utterances are produced by American speakers of eight major dialects reading phonetically rich sentences. It contains 6300 utternaces (5.4 hours) with 10 sentences from each of 630 speakers. The data preprocessing follows prior works. 
We extract spectrogram features (with a frame shift size of 10ms) from the audio, 
and the segments of 200ms duration (20 frames) are sampled from the utterances which are treated as independent sequences for learning.

\subsection{Model architecture}
\label{ss:network_architecture}
Each frame is passed to an encoder first to extract an abstract feature as the input for LSTM. Such encoder is a convolutional neural network with 5 layers of channels [64, 128, 256, 512, 128]. Every layer uses a kernel with size 4 followed by BatchNorm and LeakyReLU (except the last layer using Tanh). The dimensionality of each abstract feature is 128. The decoder is also a convolutional neural network, with 5 hidden layers of channels [512, 256, 128, 128, 64]. Each layer is followed by BatchNorm and ReLU. The whole architecture is consistent with S3VAE. For the audio dataset, we use the same architecture as DSVAE. A bi-LSTM with hidden size 256 takes the 128d features as inputs to produce $\mu$ and $\sigma$ for $s$ with size $d_s$ based on the last cell state. Another uni-LSTM produces $\mu$ and $\sigma$ for $z_{1:T}$ at each time step with dimension $d_m$ each. We set $d_s=256, d_m=32$ in Sprites, $d_s=128, d_m=8$ in MUG, $d_s=256, d_m=32$ in SM-MNIST, $d_s=256, d_m=64$ in TIMIT. $\alpha=0.9$ or $1.0$ can produce good results for all datasets. $\beta=1.0, 0.5, 5.0, 1.0$ are set respectively for Sprites, MUG, SM-MNIST, TIMIT. The dynamic prior is parameterized by an LSTM with hidden size 256.

The learning rate is set to be 0.001 for MUG, SM-MNIST and TIMIT. Sprites takes a learning rate 0.0015. C-DSVAE is trained with up to 1000 epochs (typically much fewer) on one NVIDIA V100 GPU.

\section{Supplementary Experiments}
\label{ss:additional_expts}

\begin{figure}[h!]
  \begin{subfigure}{0.495\textwidth}
    \includegraphics[width=\textwidth]{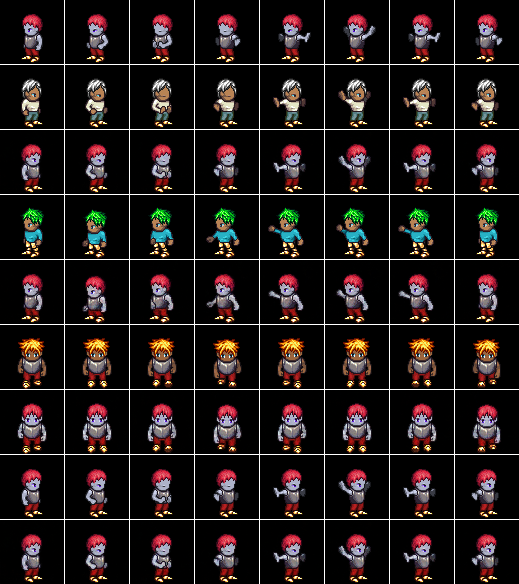}
  \end{subfigure}
  \hfill
  \begin{subfigure}{0.495\textwidth}
    \includegraphics[width=\textwidth]{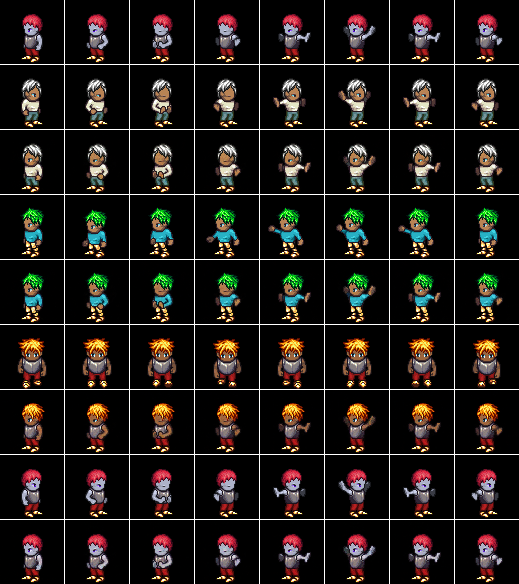}
  \end{subfigure}
  \caption{Row 1, 2, 4, 6, 8 are the raw test input sequences. \textbf{Left:} For row $i$ ($i$ is odd and $i>1$), $s$ is set to be the same as the $s$ from row 1, while $z_{1:T}$ of each row is retained. \textbf{Right:} For row $i$ ($i$ is odd and $i>1$), $z_{1:T}$ is set to be the same as the $z_{1:T}$ from row 1, while $s$ of each row is retained.}
  \label{fig:sprites_fix}
\end{figure}

\begin{figure}[h!]
  \begin{subfigure}{0.495\textwidth}
    \includegraphics[width=\textwidth]{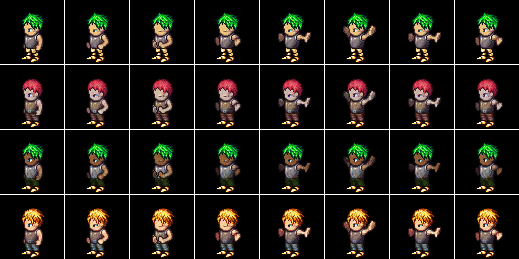}
  \end{subfigure}
  \hfill
  \begin{subfigure}{0.495\textwidth}
    \includegraphics[width=\textwidth]{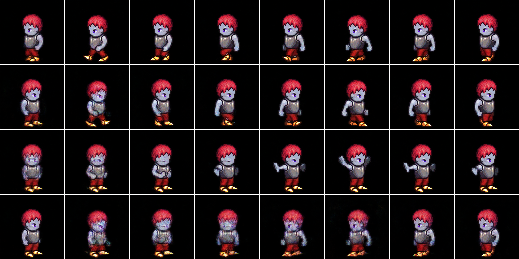}
  \end{subfigure}
  \caption{The random generations of contents and motions. \textbf{Left:} $z_{1:T}$ from row 1 is fixed and $s$ is sampled from $p(s)$ for other rows. \textbf{Right:} $s$ from row 1 is fixed and $z_{1:T}$ is sampled from $p(z_{1:T})$ for other rows.}
  \label{fig:sprites_gen}
\end{figure}

\subsection{Sprites}

We show more experiments of fixing one factor and replacing the other on Sprites in Figure~\ref{fig:sprites_fix}. In addition, Figure~\ref{fig:sprites_gen} shows the generated sequences when the factors are sampled from the priors (either static or dynamic factors). 

\subsection{MUG}

\begin{figure}[h!]
  \begin{subfigure}{\textwidth}
    \includegraphics[width=\textwidth]{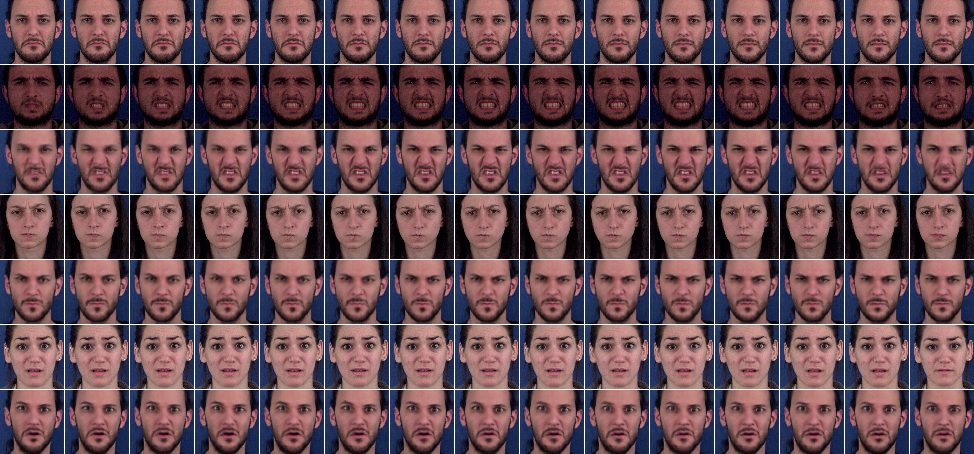}
    \caption{Fix the content. Row 1, 3, 5, 7 use the same $s$. Meanwhile, they also take $z_{1:T}$ from the previous row for generation. The person identity is well-preserved with different expressions.}
    \label{fig:mug_fix_content}
  \end{subfigure}
  ~\\~\\
  \begin{subfigure}{\textwidth}
    \includegraphics[width=\textwidth]{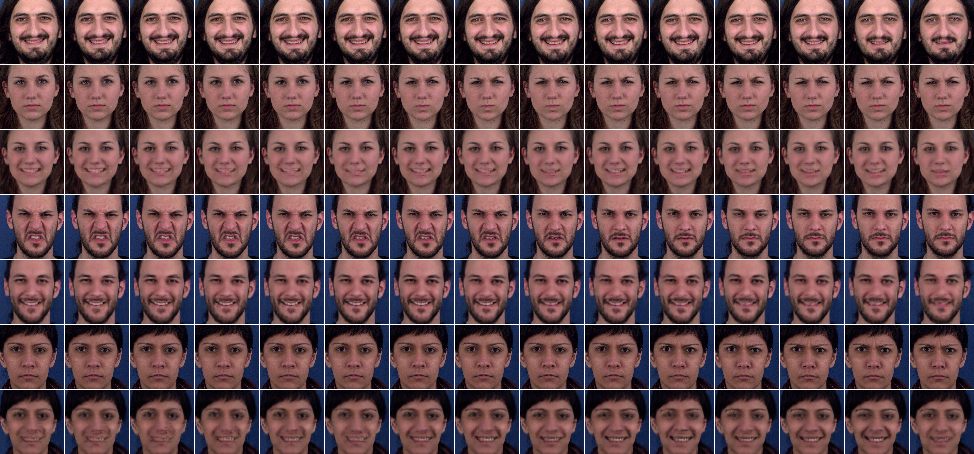}
    \caption{Fix the motion. Row 1, 3, 5, 7 use the same $z_{1:T}$. Meanwhile, they also take $s$ from the previous row for generation. All the faces have the same expression and the identity is consistent with the previous row.}
    \label{fig:mug_fix_motion}
  \end{subfigure}
  \caption{Fix either the static or dynamic factors and replace the other.}
  \label{fig:mug_fix}
\end{figure}

Figure~\ref{fig:mug_fix} shows the generated sequences when we fix either the motion or content factors. In Figure~\ref{fig:mug_fix_content}, the person identity is consistent with row 1 but has the same expression as the previous row. In Figure~\ref{fig:mug_fix_motion}, the person identity is consistent with the previous row but all with the same expression as row 1. 

\subsection{SM-MNIST}

\begin{figure}[h!]
\includegraphics[width=\textwidth]{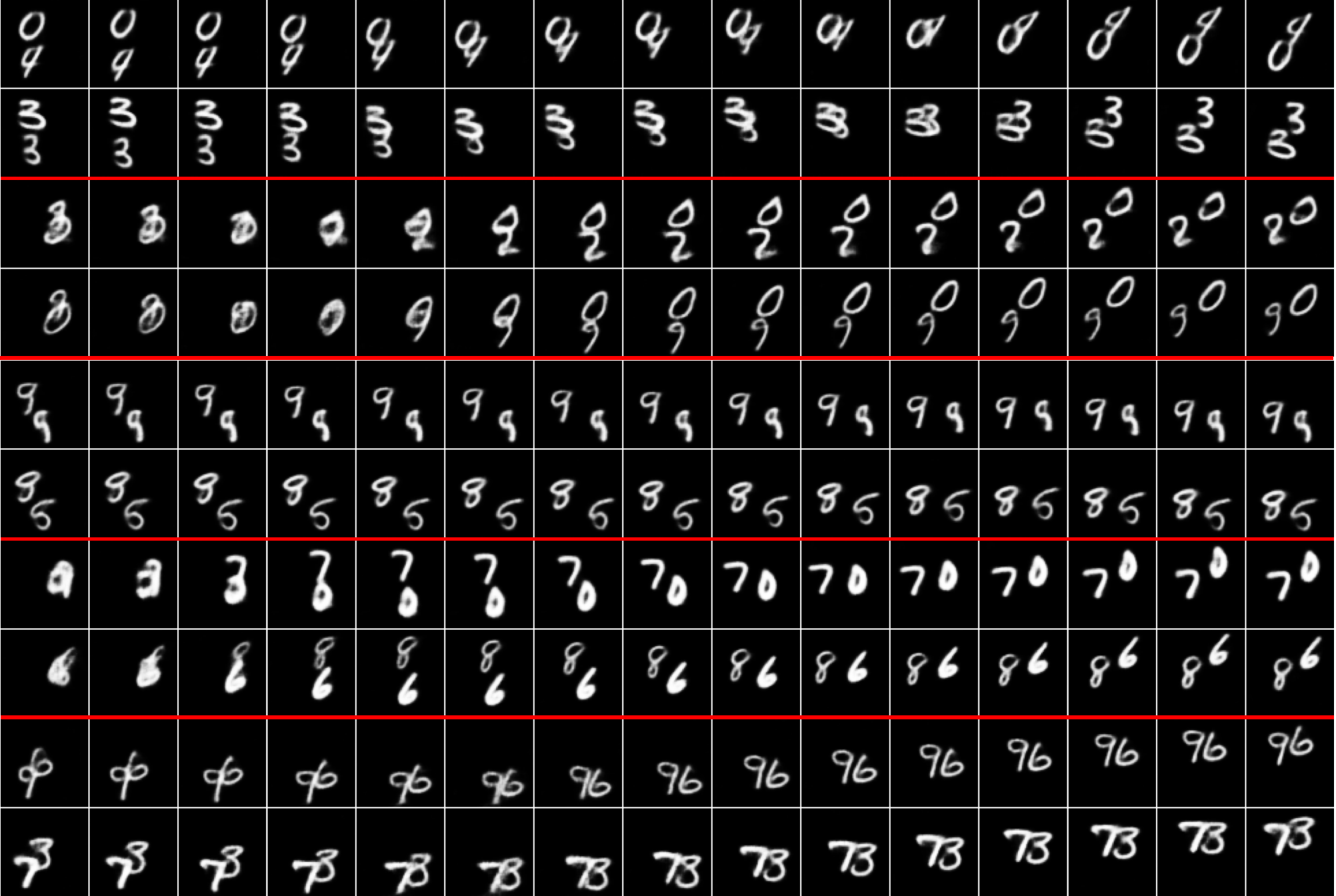}
\caption{Generate different contents. For every two rows separated by the red lines, the first row is from the test set and the second row inherits the dynamic factors $z_{1:T}$ from the first row, while samples $s$ from prior $p(s)$. As a result, the same motion is preserved with different digits.}
\label{fig:smmnist_gen_content}
\end{figure}

\begin{figure}[h!]
\includegraphics[width=\textwidth]{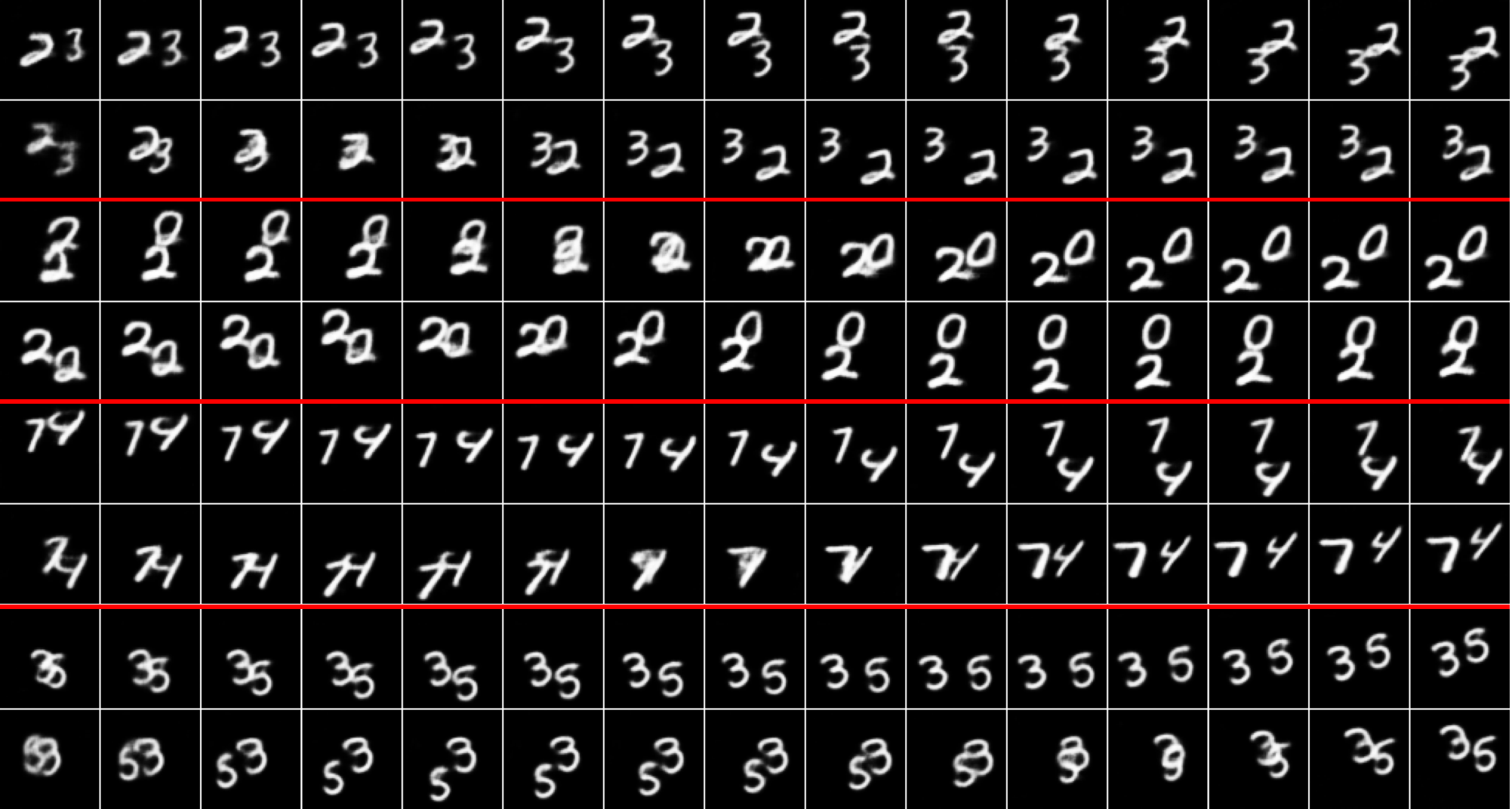}
\caption{Generate different motions. For every two rows separated by the red lines, the first row is from the test set and the second row inherits the static factor $s$. $z_{1:T}$ is sampled from the prior $p(z_{1:T})$. As a result, the same digits are preserved with different movements.}
\label{fig:smmnist_gen_motion}
\end{figure}

\begin{figure}[h!]
\includegraphics[width=\textwidth]{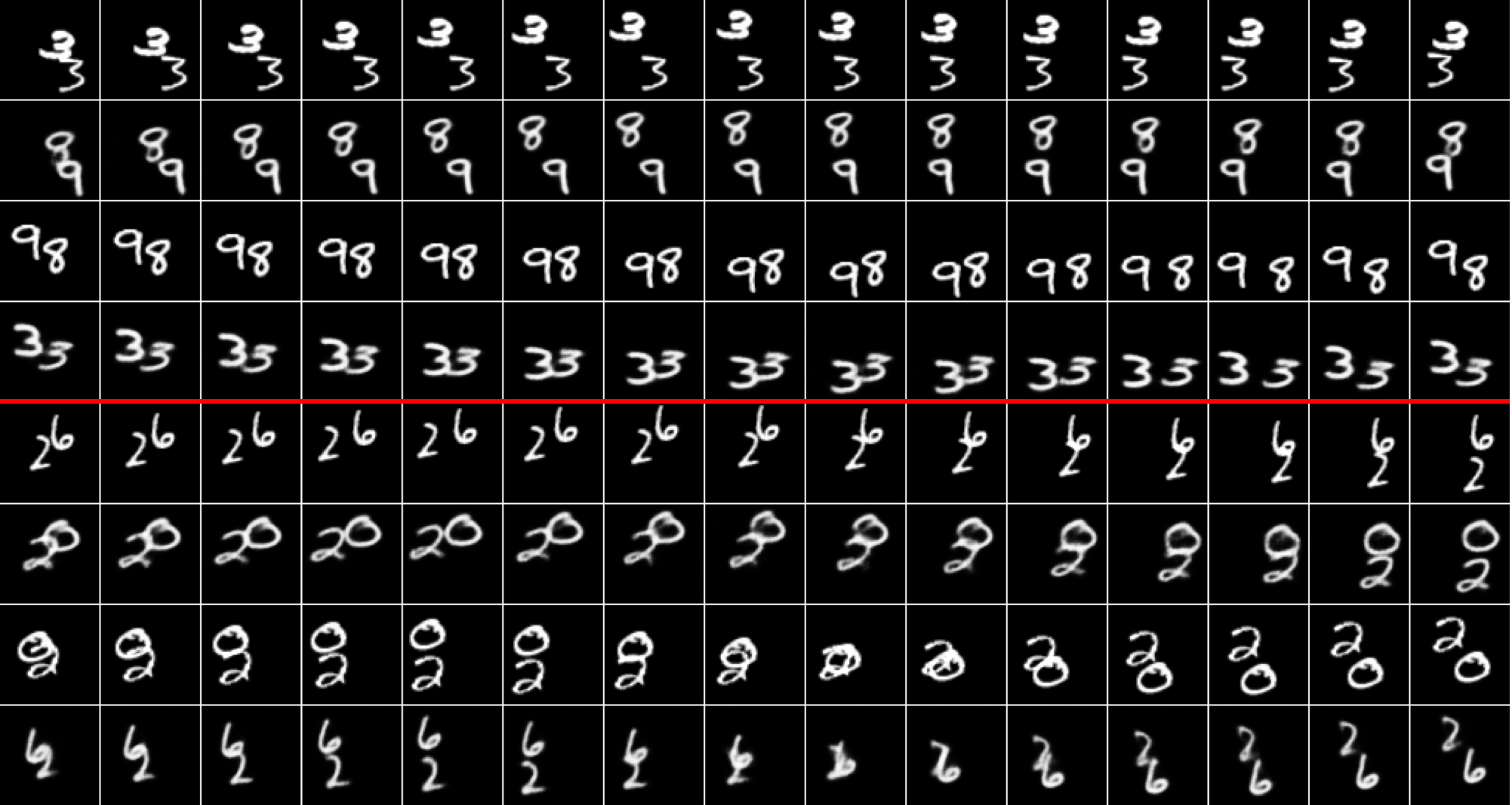}
\caption{Swap the static and dynamic factors. For every 4 rows separated by the red line, row 1 and row 3 are the raw test sequences with different contents and motions. Row 2 takes $s$ from row 3 and $z_{1:T}$ from row 1. Row 4 takes $s$ from row 1 and $z_{1:T}$ from row 3. We present 2 such swapping sets: row 1$\sim$4, row 5$\sim$8. }
\label{fig:smmnist_swap}
\end{figure}

\begin{figure}[h!]
\centering
\includegraphics[width=0.25\linewidth]{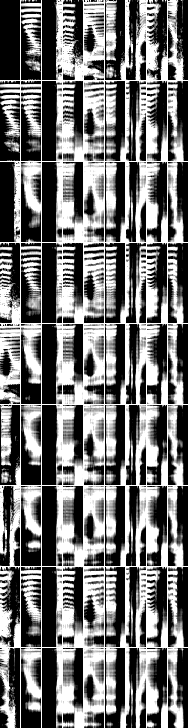}
\caption{The cross generation of 8 audio clips. Plot at $(i+1)$-th row and $(j+1)$-th column reconstructs the $i$-th static factor and $j$-th dynamic factors.}
\label{fig:timit_swap}
\end{figure}

Figure~\ref{fig:smmnist_gen_content} generates random digits from $p(s)$ to replace the content from the raw test input sequences. The motions are well-preserved while the contents are totally different. 

On the other hand, Figure~\ref{fig:smmnist_gen_motion} preserves the content digits and randomly sample $p(z_{1:T})$. The motions of the digits then become different. 

Another interesting generation task is swapping, as shown in Figure~\ref{fig:smmnist_swap}. Given two input sequences, we exchange their dynamic and static factors. 

\subsection{TIMIT}

Figure~\ref{fig:timit_swap} shows a cross-reconstruction of different audio clips. Each heatmap if of dimension $80\times 20$ which corresponds to an audio clip of length 200ms. The 80d feature at each time step is the mel-scale filter bank feature. The static factor from the first row and the dynamic factors from the first column are mixed for the generation. As we can observe, the linguistic contents are kept the same along each column and the timbres reflected as the harmonics are mostly preserved along each row.

\section{Observations on mutual information (MI)}
\label{ss:mi_observations}

\begin{figure}[h!]
  \begin{subfigure}{0.45\textwidth}
    \includegraphics[width=\textwidth]{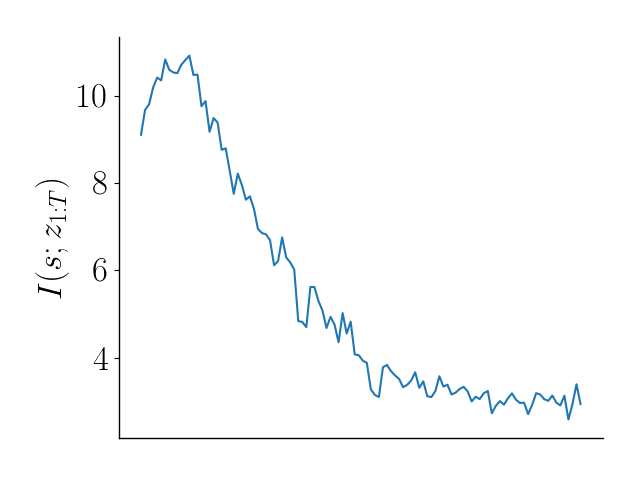}
    \caption{Sprites}
    \label{fig:mi_fz_sprites}
  \end{subfigure}
  \hfill
  \begin{subfigure}{0.45\textwidth}
    \includegraphics[width=\textwidth]{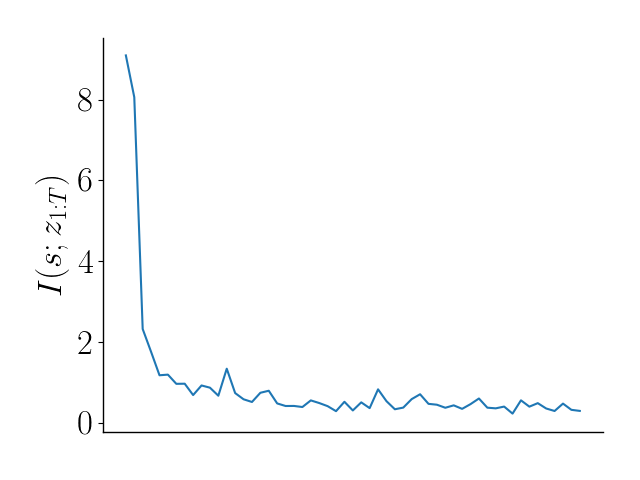}
    \caption{TIMIT}
    \label{fig:mi_fz_timit}
  \end{subfigure}
  \caption{The mutual information $I(s; z_{1:T})$ estimated by MWS decreases during the training even if we don't include $I(s;z_{1:T})$ in the objective.}
\end{figure}

\begin{figure}[h!]
  \begin{subfigure}{0.45\textwidth}
    \includegraphics[width=\textwidth]{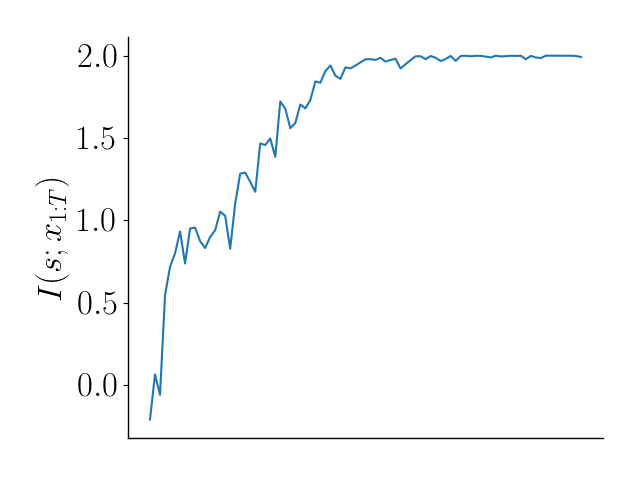}
    \caption{SM-MNIST}
  \end{subfigure}
  \hfill
  \begin{subfigure}{0.45\textwidth}
    \includegraphics[width=\textwidth]{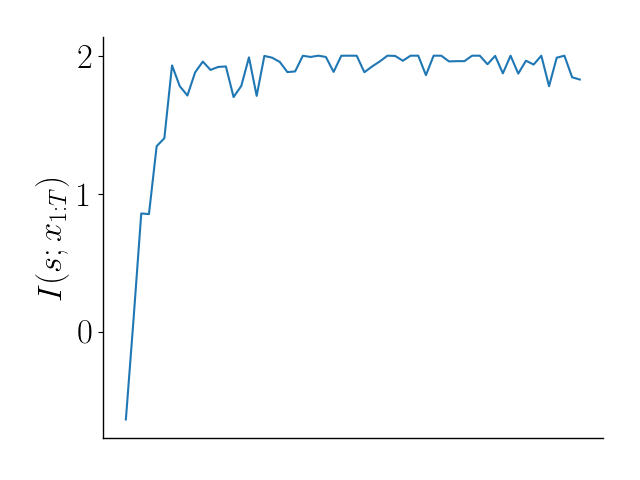}
    \caption{MUG}
  \end{subfigure}
  \caption{The mutual information $I(s, x_{1:T})$ estimated by MWS increases during the training, demonstrating the effectiveness of the contrastive estimation.}
  \label{fig:mi_xs}
\end{figure}

\subsection{MI between $s$ and $z_{1:T}$}
\label{ss:mi_sz}


Figure~\ref{fig:mi_fz_sprites} demonstrates the curve of MI $I(s;z_{1:T})$ during training on Sprites when we don't include $I(s;z_{1:T})$ in the objective function. We estimate $I(s;z_{1:T})$ using the Minibatch Weighted Sampling (MWS)  estimation (though not optimized directly). The curve increases in the early stage of  training, implying that while learning useful representations for reconstruction,  disentanglement is compromised in the early stage. But as the training continues, the model picks up the disentanglement terms besides the reconstruction. Figure~\ref{fig:mi_fz_timit} demonstrates the same experiment on another dataset TIMIT. 

\subsection{MI between $s$ and $x_{1:T}$}

We show in Figure~\ref{fig:mi_xs} that, the contrastive estimation could boost $I(s;x_{1:T})$ during training. Our training process optimizes the contrastive estimation, but as we can see in the figure, the MI term $I(s;x_{1:T})$ estimated by MWS also increases accordingly.

\section{Ablation Study on Estimation Methods}
\label{s:ablation}

\begin{figure}[!htp]
\centering
\includegraphics[width=\linewidth]{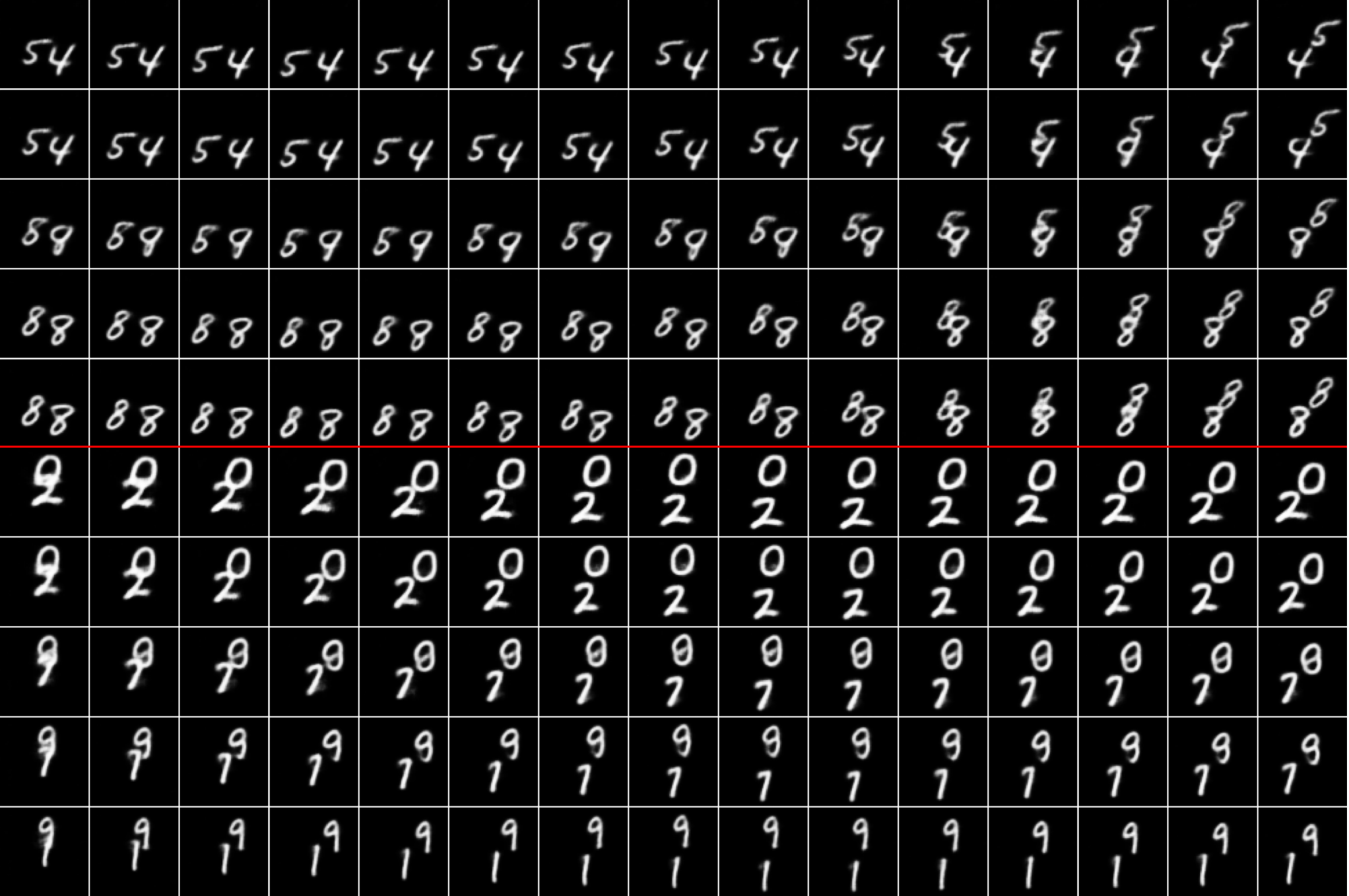}
\caption{Interpolation in the latent space. For every 5 rows separated by the red line, the first and last rows have different contents $s$, but share the same motion. The 3 rows in between keep the same motion but their $s$'s linearly interpolate between row 1 and row 5. One can observe that the content transition is smooth, while the motion is intact. }
\label{fig:smmnist_interpolate}
\end{figure}

\begin{table}[t]
\centering
\caption{Compare MWS and contrastive estimation of $I(s;x_{1:T})$ and $I(z_{1:T};x_{1:T})$ on MUG. "all MWS est" means that C-DSVAE optimizes MI terms all estimated by MWS rather than the contrastive estimation.}
\label{tab:est_mug}
\begin{tabular}{ccccc}
\toprule
Methods & Acc$\uparrow$ & IS$\uparrow$ & H(y|x)$\downarrow$ & H(y)$\uparrow$ \\
\midrule
DSVAE & 54.29\% & 3.608 & 0.374 & 1.657 \\
DSVAE+all MWS est & 66.25\% & 4.796 & 0.175 & 1.743 \\
C-DSVAE & 81.16\% & 5.341 & 0.092 & 1.775 \\
\bottomrule
\end{tabular}
\end{table}
\begin{table}[t]
\centering
\caption{Compare MWS and contrastive estimations of $I(s;x_{1:T})$ and $I(z_{1:T};x_{1:T})$ on SM-MNIST.}
\label{tab:est_smmnist}
\begin{tabular}{ccccc}
\toprule
Methods & Acc$\uparrow$ & IS$\uparrow$ & H(y|x)$\downarrow$ & H(y)$\uparrow$ \\
\midrule
DSVAE & 88.19\% & 6.210 & 0.185 & 2.011 \\
DSVAE+all MWS est & 91.81\% & 6.312 & 0.205 & 2.107 \\
C-DSVAE & 97.84\% & 7.163 & 0.145 & 2.176 \\
\bottomrule
\end{tabular}
\end{table}

\begin{table}[h!]
\centering
\caption{Performance with different batch sizes on TIMIT.}
\label{tab:bs_sensitivity}
\begin{tabular}{ccccc}
\toprule
batch size & content EER$\downarrow$ & motion EER$\uparrow$ \\
\midrule
64 & 4.21\% & 30.23\% \\
128 & 4.07\% & 31.42\% \\
256 & 4.03\% & 31.81\% \\
\bottomrule
\end{tabular}
\end{table}

Our C-DSVAE estimates $I(s;x_{1:T})$, $I(z_{1:T};x_{1:T})$ with contrastive learning and $I(s;z_{1:T})$ with MWS. To further demonstrate the advantage of the contrastive estimation, we compare it with the model with MWS estimations on all the MI terms including $I(s;x_{1:T})$, $I(z_{1:T};x_{1:T})$. Table~\ref{tab:est_mug} and ~\ref{tab:est_smmnist} give the observations. With all MWS estimations, the performance would drop heavily.

These results help support the claim that the inductive biases brought by contrastive learning might contribute more to the good performance than the MI estimation. Also note that contrastive learning works well when the goal of learning is to maintain some invariance between different views. For other MI estimation tasks, contrastive learning might not be the best option.

\section{Interpolation in Latent Space}
\label{s:latent_traversal}

To further show our learnt latent space is smooth and meaningful, we linearly interpolate between 2 content factors corresponding to different digit pairs and generate the sequences. In Figure~\ref{fig:smmnist_interpolate}, one can see that the transition from one content to another is smooth. In the first example, ("5", "4") gradually changes to ("8", "8"). In the second example, ("0", "2") gradually changes to ("9", "1"). While "2" transforms to "1", it first becomes "7" and its font gets slimmer.

\section{Sensitivity Analysis on Batch Size}

In Table~\ref{tab:bs_sensitivity}, we show how different batch sizes would affect the evaluation performance on TIMIT. Batch size 256 gives the best numbers.

\end{document}